\documentclass[lettersize,journal]{IEEEtran}

\usepackage{amsmath,epsfig}

\usepackage{caption}
\usepackage{enumitem}
\setlist[itemize]{noitemsep,nolistsep}

\let\OLDthebibliography\thebibliography
\renewcommand\thebibliography[1]{
  \OLDthebibliography{#1}
  \setlength{\parskip}{0pt}
  \setlength{\itemsep}{0pt plus 0.3ex}
}

\usepackage{pifont}% http://ctan.org/pkg/pifont

\newcommand{\cmark}{\ding{51}}%
\usepackage{graphicx}
\usepackage{amsmath}
\usepackage{amssymb}
\usepackage{booktabs}

\usepackage{amsmath,amsfonts}
\usepackage{algorithmic}
\usepackage{algorithm}
\usepackage{array}
\usepackage{textcomp}
\usepackage{stfloats}
\usepackage{url}
\usepackage{verbatim}
\usepackage{graphicx}
\usepackage{cite}
\usepackage{xspace}
\makeatletter
\DeclareRobustCommand\onedot{\futurelet\@let@token\@onedot}
\def\@onedot{\ifx\@let@token.\else.\null\fi\xspace}

\makeatother

\usepackage{amsmath}
\usepackage{amssymb}
\usepackage[table]{xcolor}
\usepackage[pagebackref=true,breaklinks=true,colorlinks=true,bookmarks=false,linkcolor={red!50!black},urlcolor={darkgray!80!black},citecolor={green!50!black}]{hyperref}
\usepackage[group-separator={,}]{siunitx}
\usepackage{multirow}
\usepackage{tabularx}
\usepackage{booktabs} % To thicken table lines
\usepackage{makecell}
\usepackage{graphbox}
\usepackage{footmisc}
\usepackage{bm}
\usepackage{caption}
\usepackage{arydshln}
\usepackage{dashrule}
\usepackage{subcaption}
\usepackage{enumitem}
\setlist[itemize]{noitemsep,nolistsep}

\usepackage[capitalize]{cleveref}
\crefname{section}{Sec.}{Secs.}
\Crefname{section}{Section}{Sections}
\Crefname{table}{Table}{Tables}
\crefname{table}{Tab.}{Tabs.}
\Crefname{figure}{Figure}{Figures}
\crefname{figure}{Fig.}{Figs.}
\Crefname{equation}{Equation}{Equations}
\crefname{equation}{Eq.}{Eqs.}

\usepackage{graphicx}
\usepackage{amsmath}
\usepackage{amssymb}
\usepackage{booktabs}

\usepackage{graphicx}
\usepackage{makecell}
\usepackage{comment}
\usepackage{amsmath,amssymb} % define this before the line numbering.
\usepackage{color}
\usepackage{booktabs}
\usepackage{arydshln}
\usepackage{multirow}
\usepackage{hyperref}

\usepackage[]{lineno}

\renewcommand{\paragraph}[1]{ 
 \noindent\textbf{#1}~}

\newcommand{\blue}[1]{\textbf{\textcolor{mblue}{#1}}}

\newcommand{\red}[1]{\textcolor{red}{#1}}
\newcommand{\bred}[1]{\textbf{\textcolor{red}{#1}}}
\newcommand{\green}[1]{\textcolor{mgreen}{#1}}

\newcommand{\gray}[1]{\textcolor{mgray}{#1}}

\colorlet{lightcyan}{cyan!10}
\colorlet{lightpink}{pink!20}
\colorlet{lightgray}{gray!10}

\definecolor{mgray}{gray}{0.35}
\definecolor{mred}{RGB}{238, 34, 12}
\definecolor{mgreen}{RGB}{1, 127, 0}
\definecolor{mblue}{RGB}{0, 77, 128}
\definecolor{orange}{RGB}{240, 120,0}

\usepackage[table]{xcolor}
\usepackage[pagebackref=true,breaklinks=true,colorlinks=true,bookmarks=false]{hyperref}

\usepackage[capitalize]{cleveref}
\crefname{section}{Sec.}{Secs.}
\Crefname{section}{Section}{Sections}
\Crefname{table}{Table}{Tables}
\crefname{table}{Tab.}{Tabs.}

\begin{document}

% Example definitions.
% --------------------
\def\x{{\mathbf x}}
\def\L{{\cal L}}

% Title.
% ------
\title{Towards Robust Text-Prompted Semantic Criterion for In-the-Wild Video Quality Assessment}
%
% Single address.
% ---------------
\author{Haoning Wu,
Liang Liao,~\IEEEmembership{Member,~IEEE},
Annan Wang,
        Chaofeng Chen,
        Jingwen Hou,~\IEEEmembership{Student Member,~IEEE},
        \\
        Wenxiu Sun,
        Qiong Yan,
        Weisi Lin,~\IEEEmembership{Fellow,~IEEE}

\thanks{H. Wu, L. Liao, A. Wang and C. Chen are with the S-Lab, Nanyang Technological University, Singapore; J. Hou, and W. Lin are with the School of Computer Science and Engineering, Nanyang Technological University, Singapore. (e-mail: haoning001@e.ntu.edu.sg; liang.liao@ntu.edu.sg; c190190@e.ntu.edu.sg; chaofeng.chen@ntu.edu.sg; jingwen003@e.ntu.edu.sg; wslin@ntu.edu.sg;)}
\thanks{W. Sun and Q. Yan are with the Sensetime Research, Hong Kong, China. (e-mail: irene.wenxiu.sun@gmail.com;  sophie.yanqiong@gmail.com)}
\thanks{Corresponding author: Weisi Lin.}}
%\name{Anonymous ICME submission}
%Address and e-mail should NOT be added in the submission paper. They should be present only in the camera ready paper. 

\maketitle

\begin{abstract}
The proliferation of videos collected during in-the-wild natural settings has pushed the development of effective Video Quality Assessment (VQA) methodologies. Contemporary supervised opinion-driven VQA strategies predominantly hinge on training from expensive human annotations for quality scores, which limited the scale and distribution of VQA datasets and consequently led to unsatisfactory generalization capacity of methods driven by these data. On the other hand, although several handcrafted zero-shot quality indices do not require training from human opinions, they are unable to account for the semantics of videos, rendering them ineffective in comprehending complex authentic distortions (e.g., white balance, exposure) and assessing the quality of semantic content within videos. To address these challenges, we introduce the text-prompted Semantic Affinity Quality Index (\underline{SAQI}) and its localized version (SAQI-Local) using Contrastive Language-Image Pre-training (CLIP) to ascertain the affinity between textual prompts and visual features, facilitating a comprehensive examination of semantic quality concerns without the reliance on human quality annotations. By amalgamating SAQI with existing low-level metrics, we propose the unified Blind Video Quality Index (\underline{BVQI}) and its improved version, BVQI-Local, which demonstrates unprecedented performance, surpassing existing zero-shot indices by at least 24\% on all datasets. Moreover, we devise an efficient fine-tuning scheme for BVQI-Local that jointly optimizes text prompts and final fusion weights, resulting in state-of-the-art performance and superior generalization ability in comparison to prevalent opinion-driven VQA methods. We conduct comprehensive analyses to investigate different quality concerns of distinct indices, demonstrating the effectiveness and rationality of our design. Our code is accessible at \url{https://github.com/VQAssessment/BVQI}.
\end{abstract}
%
%\begin{keywords}
%Opinion-unaware Video Quality Assessment, Semantic Affinity, Technical Metrics
%\end{keywords}
%
\section{Introduction}
\label{sec:intro}

\IEEEPARstart{W}{ith} the exponential growth of online videos, there has been an increased interest among researchers and the industry in the field of video quality assessment (VQA), to evaluate, recommend, and potentially enhance the quality of immense volume of videos captured by users in the wild. Compared with traditional VQA tasks~\cite{livevqa,csiqvqa}, in-the-wild VQA is much more difficult as real-world videos can suffer from complicated and various quality degradations (\textit{e.g. out-of-focus, motion blur, bad white balance, noise, over/under-exposure}) and do not have pristine counterparts as references.

\begin{figure}
    \centering
\includegraphics[width=\linewidth]{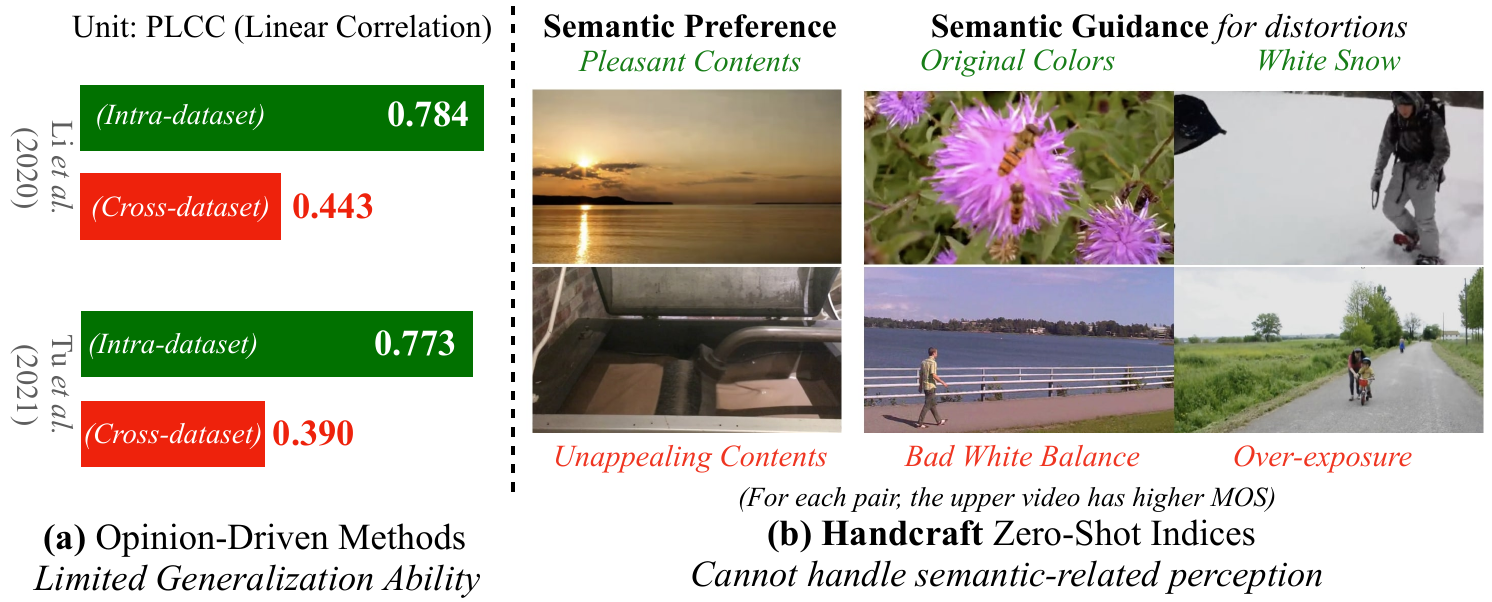}
    %\vspace{-9pt}
    \caption{\textbf{(a)} Due to the limited scale of VQA datasets, opinion-driven VQA methods usually suffer from limited generalization ability; \textbf{(b)} Existing zero-shot quality indices are based on low-level handcraft features, failing to handle semantic-related perception during quality evaluation.}
    \label{fig:challenge}
    \vspace{-18pt}
\end{figure}

In past years, opinion-driven VQA approaches~\cite{tlvqm,fastvqa,vsfa,gstvqa,cnntlvqm,videval,rirnet} have been extensively studied and have achieved significant performance improvements. However, they rely heavily on human opinions to make the model fit the data distribution in the training datasets~\cite{pvq,cvd,kv1k,vqc}, which presents a significant challenge. Specifically, collecting large-scale human opinions is a costly process, requiring efforts from at least 15 (\textit{often more than 100\cite{kv1k,ytugccc}}) annotators~\cite{itu} to obtain reliable mean opinion scores (MOS) for each video. %With a consequentially limited scale of training datasets, opinion-driven VQA methods usually suffer from limited generalization ability
As a result, training datasets for opinion-driven VQA methods are often limited in scale, leading to their limited generalization ability on new datasets. For instance, VQA methods trained sorely with KoNViD-1k\cite{kv1k} (1200 labelled videos) can only poorly correlate (as in Fig.~\ref{fig:challenge}(a)) with human opinions in YouTube-UGC\cite{ytugccc} (1380 labelled videos). The limited and unstable generalization performance due to the limited dataset scale severely challenges the practical usability of opinion-driven VQA methods.
%all methods can achieve no more than 0.35 PLCC on YouTube-UGC dataset while training with LIVE-VQC dataset.

%they rely on large amounts of training data with expensive human subjective scores and are typically not easily adaptable to new datasets. How to alleviate the burden of costly training data and build a robust VQA capable of evaluating any given video is the issue that urgent to study.

% In contrast, opinion-unaware VQA methods do not require expensive training data and tend to exhibit good robustness across different datasets, motivating us to explore on these methods.

\begin{figure}
    \centering
\includegraphics[width=\linewidth]{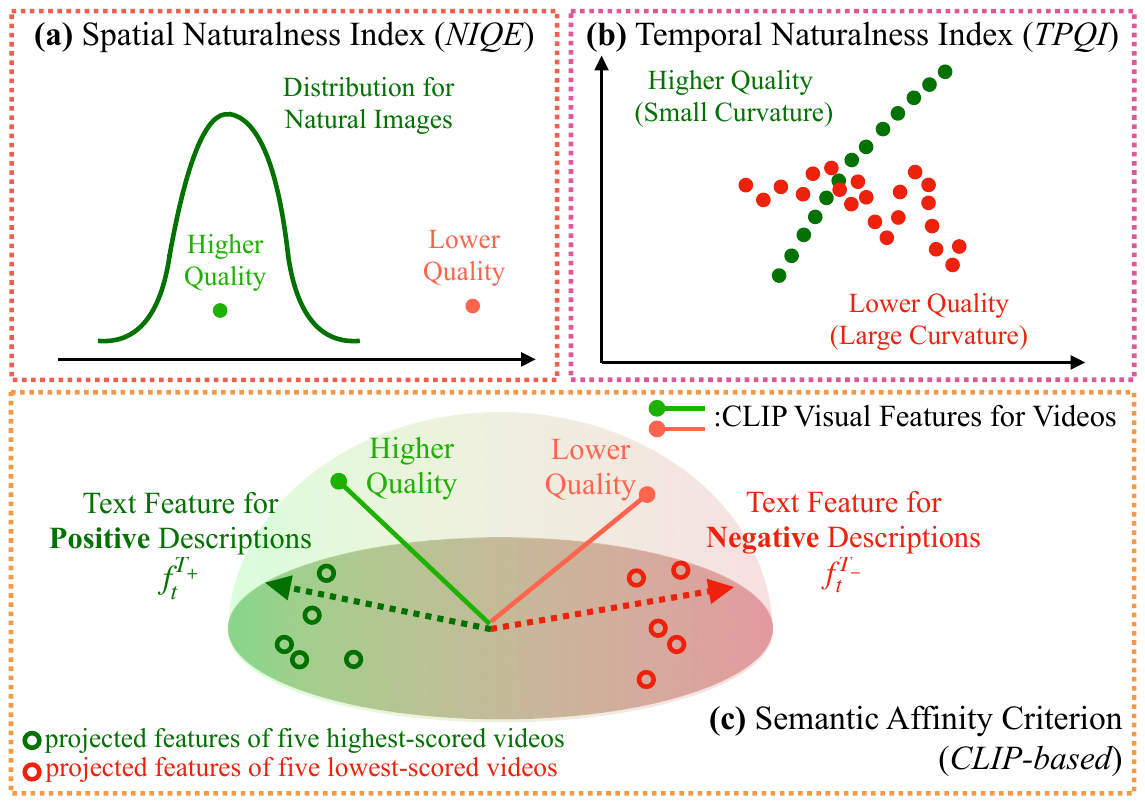}
    %\vspace{-9pt}
    \caption{Criteria for handcraft spatial \textbf{(a)} and temporal \textbf{(b)} naturalness indices, and the proposed \textbf{(c)} semantic affinity criterion, which can well distinguish the best (\green{green circle}) and worst (\red{red circle}) videos in~\cite{ytugc} without regression.}
    \label{fig:criterion}
    \vspace{-8pt}
\end{figure}

%现有的opinion-unaware还不能解决什么问题
%authentic distortion; aesthetics
The challenge motivates us to explore zero-shot VQA approaches that do not rely on expensive human annotations for video quality scores. For example, NIQE~\cite{niqe} measures \textit{spatial} naturalness of images by comparing their Multivariate Gaussian (MVG) distributions with those of pristine natural contents (Fig.~\ref{fig:criterion}(a)). TPQI~\cite{tpqi}, inspired by knowledge of the human visual system, measures the \textit{temporal} naturalness of videos through the inter-frame curvature on perceptual domains~\cite{primaryv1,lgn}. Although these metrics have proven to work well under traditional low-level distortions (e.g. \textit{compression artifacts}), they still perform poorly~\cite{ytugccc,videval} for in-the-wild VQA as they are not aware of semantic information in videos. As semantic information might directly affect the quality score of a video, observed as \textbf{\textit{semantic preference}} (Fig.~\ref{fig:challenge}(b) \textit{left}) by ~\cite{vsfa,dover,rapique}), or provide \textbf{\textit{semantic guidance}}    (Fig.~\ref{fig:challenge}(b) \textit{right})~\cite{dbcnn,spaq,rfugc} to understand \textit{authentic distortions} with similar low-level patterns to non-distorted situations, these handcraft indices are not enough for in-the-wild VQA. Thus, it is crucial to design a robust semantic-aware criterion that does not require costly human-annotated quality scores for training.

%Moreover, many existing studies have noticed that natural authentic distortions~\cite{spaq,paq2piq,ytugc} or aesthetic-related issues~\cite{vsfa,dover} commonly occur on in-the-wild videos and impact human quality perception. These issues are hardly captured with these low-level criteria, but could be better extracted with semantic-aware deep neural features\cite{cnntlvqm,fastervqa,videval,dbcnn,bvqa2021,mdtvsfa}.

%(Fig.~\ref{fig:criterion}(a)/(b)), \cfchen{including spatial naturalness prior cite[NIQE] and temporal naturalness prior cite[TPQI] as shown in fig(a)(b). The former one measures distance between xxx and xxx, it mainly consider authentic $\ldots$. The latter evaluates the temporal perceptual straightness. Although they achieve great performance, most of them does not consider semantic xxx. Meanwhile, high level \textbf{aesthetic-related} quality $\ldots$.} without considering  semantics of videos. Henceforth, \textbf{authentic distortions}\cite{spaq,paq2piq,ytugc} such as\textit{over/under exposure, out-of-focus, bad colors} that commonly occur on in-the-wild videos are difficult to be captured\cite{videval} by these indices. Furthermore, higher-level \textbf{aesthetic-related} quality issues~\cite{dover,vsfa}, such as \textit{hard-to-understand contents} or \textit{chaotic compositions}, also significantly impact the quality of in-the-wild videos, yet are even harder to be measured by these semantic-unaware naturalness metrics.

To evaluate the video quality at a higher semantic level, we propose using the Contrastive Language-Image Pre-training (CLIP)\cite{clip}, a vision-language model pre-trained on a massive number (\textit{about 400 million}) of naturally existing image-text pairs from the internet, which provides more robust semantic understanding in real-world scenarios. More importantly, CLIP proves to robustly measure the affinity between visual inputs and diverse textual inputs. Based on this, we propose a CLIP-based semantic affinity criterion that evaluates videos based on their affinity to positive and negative text descriptions of quality. Specifically, the criterion defines that videos with higher affinity to positive text descriptions of quality shall have higher quality than those with more similarity to negative descriptions. We introduce the Semantic Affinity Quality Index (\textbf{SAQI}) that uses this criterion to measure video quality in a zero-shot manner, without any human quality labels for training. Moreover, the SAQI evaluates both aesthetic and authentic distortions in videos using two different positive-negative description pairs to account for semantic-related challenges as shown in Fig.~\ref{fig:challenge}(b). It measures semantic preference (goodness of contents) using the \textit{good$\leftrightarrow$bad} pair and evaluates semantic-guided authentic distortions using the \textit{high$\leftrightarrow$low quality} pair, resulting in consistent and effective performance.

Contrary to handcrafted indices, deep learning-based SAQI is less sensitive to low-level textures~\cite{videval} and incapable of measuring temporal quality. Therefore, it can synergize with existing spatial and temporal naturalness indices and be integrated into a more powerful video quality index that does not rely on human opinions. To achieve this, we introduce a Gaussian normalization followed by a sigmoid rescaling process~\cite{vqeg} to align the scales between the raw low-level metrics and the proposed SAQI. Once aligned, the indices can be combined into the unified Blind Video Quality Index (\textbf{BVQI}\footnote{The BVQI is previously named as the BUONA-VISTA (\textit{abbr.} for Blind Unified Opinion-Unaware Video Quality Index via Semantic and Technical Aggregation) in conference version~\cite{buonavista} and shortened to facilitate reading.}).

%Existing opinion-unaware VQA methods~\cite{niqe,ilniqe,brisque,tpqi} typically consider low-level statistic information of videos, which can detect several types of traditional distortions, \textit{e.g. JPEG artifacts, synthetic noises or transmission errors}. However, many subjective and objective studies have demonstrated that in-the-wild videos also contain \textbf{authentic distortions}~\cite{spaq,paq2piq,ytugc}, e.g. \textit{over/under exposure, out-of-focus, bad color, or motion distortions}. These authentic distortions are usually hard to be captured by existing opinion-unaware indices~\cite{videval}. Moreover, several studies~\cite{dover,vsfa} have noticed that higher-level \textbf{aesthetic-related} quality issues, such as \textit{hard-to-understand contents, chaotic compositions} also affect the quality of in-the-wild videos, which are also difficult to be measured by traditional metrics.

% Semantic 

%While they are able to accurately measure the the statistical naturalness of a video, they usually lack the ability to understand the semantic content of the video, nor to distinguish between the aesthetic-related quality perception, \textit{e.g.} the meaningfulness of videos. Consequently, these methods are usually with very bad performance when they try to evaluate on non-natural videos, \textit{e.g.} in the YouTube-UGC dataset. 

This paper significantly expands upon our previously-proposed method~\cite{buonavista}, which presented two main contributions. Firstly, it introduced the CLIP-based SAQI for zero-shot VQA, which sufficiently matches human quality perception by incorporating antonym-differential affinity and multi-prompt aggregation. Secondly, it introduced Gaussian normalization and sigmoid rescaling strategies to align and aggregate the proposed SAQI with low-level technical metrics into comprehensive \textbf{BVQI} (or \textbf{BUONA-VISTA}, as in original version) quality index, which outperforms existing zero-shot VQA indices by \textbf{\textit{at least 20\%}} on all datasets. In this extension, we present three additional substantial improvements:

\begin{enumerate}
    \item We propose a localized semantic affinity quality index (\textbf{SAQI-Local}) via modifying the attention pooling layer in the CLIP model, and a respective improved version of BVQI, \textbf{BVQI-Local}, which not only achieves higher accuracy for zero-shot VQA but also enables a robust and flexible semantic-aware quality localizer.
    \item We propose an efficient fine-tuning scheme for BVQI-Local, which can achieve state-of-the-art performance among training-based VQA methods (\textbf{18\%} better than the zero-shot version), with only a few parameters to be optimized. The fine-tuned version also proves much better robustness than existing methods.
    \item We conduct extensive analyses, including local quality maps, evaluation on more concrete prompts, and analysis on downsampling, which provide strong evidence that the proposed SAQI improves in-the-wild VQA by focusing on the aforementioned semantic concerns.
\end{enumerate}

\section{Related Works}

\subsection{No-reference Video Quality Assessment}

 Unlike full-reference VQA, no-reference VQA can only predict quality based on features from the distorted videos. Classically, several approaches \cite{brisque,bofqa,rrstedqa,diivine,stgreed} employ handcrafted features to evaluate video quality without references. Some methods \cite{niqe,viideo,vbliinds,tpqi} hypothesize that they can predict quality scores from statistical hypotheses without regression from any subjective human opinions (\textit{i.e.} annotations), usually categorized as {opinion-unaware} or {completely blind} video quality indices. On the contrary, some other methods \cite{tlvqm,rapique,videval} choose to first handcraft quality-sensitive features and then regress them to human-labelled subjective mean opinion scores (MOS), in order to better match human perception. With additional training data, these regression-based methods usually reach better in-distribution performance, yet they are usually less robust and predict less accurately across datasets.
 
 %Therefore, their capacity to generalize new types of content from one VQA dataset to another is substantially worse, suggesting that non-technical impacts are necessary to be considered in the UGC-VQA problem.

%\paragraph{Deep VQA Methods.} 
Recently, considering the non-negligible importance of semantics in NR-VQA, deep VQA methods \cite{fastvqa, dctqa, cnn+lstm,deepvqa, gstvqa, vsfa, mlsp, dstsvqa, svqa, mdtvsfa} with semantic pre-training are becoming predominant.  VSFA \cite{vsfa} conducts subjective studies to demonstrate videos with more attractive semantics receive higher subjective ratings. Therefore, it uses the semantic-aware features extracted by pre-trained ResNet-50 \cite{he2016residual} from ImageNet-1k dataset \cite {imagenet} and adopts Gate Recurrent Unit (GRU)~\cite{gru} for quality regression, followed by several more recent approaches~\cite{mlsp,lsctphiq,pvq,rirnet,cnntlvqm,discovqa}. Though reaching better performance, opinion-driven deep VQA methods are also facing the same problem of limited generalization ability across datasets, as there is no existing semantic-aware video quality indices which does not require training with human opinions. This motivates us to design a robust semantic-aware zero-shot quality index.

\begin{figure*}
    \centering
    \includegraphics[width=0.96\textwidth]{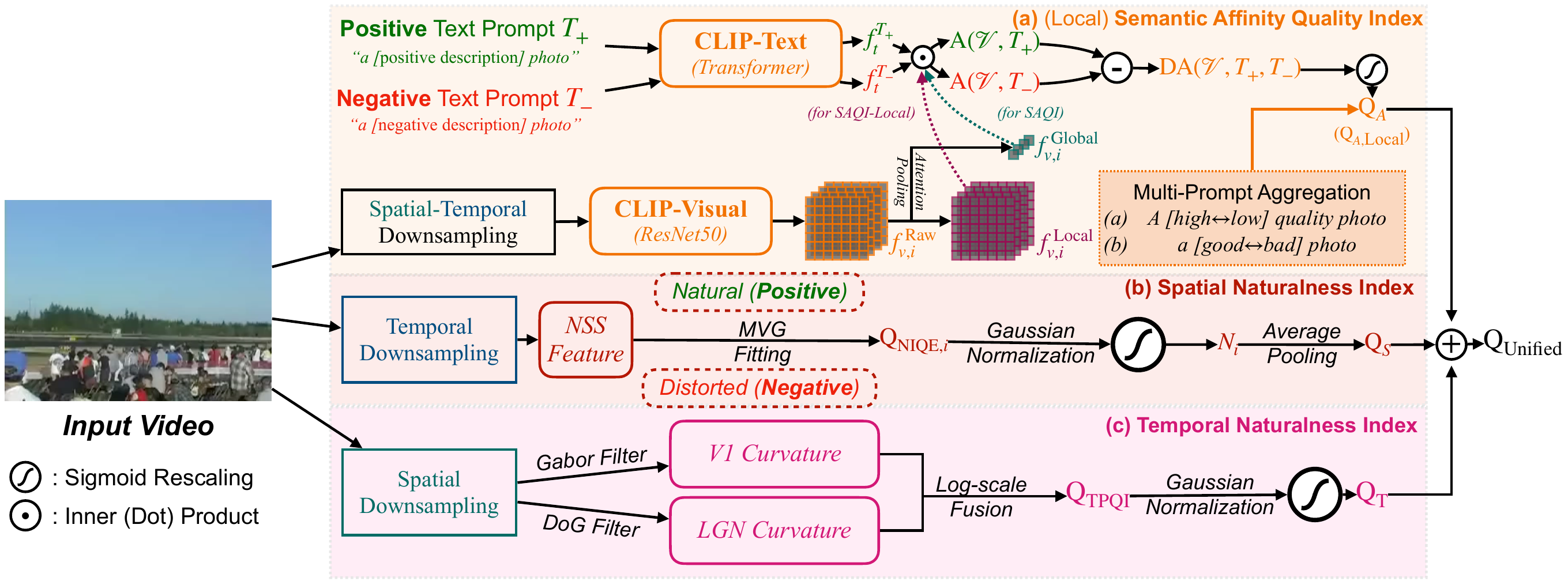} 
    %\vspace{-10bpt}
    \caption{The overall pipeline of \textbf{BVQI} and \textbf{BVQI-Local}, including \textbf{(a)} Semantic Affinity Quality Index (baseline version SAQI and localized version SAQI-Local), \textbf{(b)} Spatial Naturalness index, and \textbf{(c)} Temporal Naturalness Index. The three indices are aligned and aggregated to the final predictions.}
    \label{fig:1}
    \vspace{-10pt}
\end{figure*}

\subsection{Vision-Language Pre-training}

In recent years, vision-language models~\cite{clip,align,xclip,coca,simvlm} have emerged as a predominant type of foundation models, with the ability to learn joint representations across visual and textual information. Among them, CLIP~\cite{clip} and ALIGN~\cite{align} share a similar training paradigm to increase the affinity between paired text sentences and images, and decrease the affinity between the unpaired ones from a very large-scale paired vision-language training dataset. Unlike pure vision foundation models~\cite{he2016residual,vit} pre-trained from annotated classification datasets~\cite{imagenet,in21k}, the vision-language pre-training enables downstream tasks to measure the explicit affinity from semantic-aware deep visual features to various natural language prompts~\cite{denseclip, clipiqa}. Moreover, contrary to a few most recent studies on IQA\cite{clipiqa,liqe} attempting to perceive traditional low-level distortions with CLIP (as proved ineffective in Tab.~\ref{tab:downsample}), we design and prompt the CLIP to mainly focus on semantic goodness and global distortions (e.g. bad exposure) that need semantics to be well-understood, and leave the low-level distortion modeling through existing handcraft zero-shot quality indices, which proves better and more stable performance across different in-the-wild VQA datasets.

\section{The Proposed Zero-Shot Quality Index}

In this section, we introduce the three metrics with different criteria that make up the proposed video quality index, including the CLIP-based SAQI ($\mathrm{Q}_A$, Sec.~\ref{sec:qa}), and two technical naturalness metrics: the Spatial Naturalness Index ($\mathrm{Q}_S$, Sec.~\ref{sec:qs}), and the Temporal Naturalness Index ($\mathrm{Q}_T$, Sec.~\ref{sec:qt}). The three indices are aligned and aggregated into the proposed \textbf{BVQI} quality index. Moreover, with consideration on the locality for human quality perception, we propose the localized version of SAQI, the SAQI-Local, and its respective BVQI-Local. The overall pipeline of BVQI/BVQI-Local is illustrated in Fig.~\ref{fig:1}, discussed as follows.

\subsection{The Semantic Affinity Quality Index (SAQI, $\mathrm{Q}_A$)}
\label{sec:qa}
To evaluate semantic-related quality perception (goodness of contents, semantic-aware distinction on distortions), we design the {Semantic Affinity Quality Index} (\textbf{SAQI}, $\mathrm{Q}_A$) as follows. 

\subsubsection{Focusing on Semantics through Downsampling}
\label{sec:downsample}
As the SAQI aims at authentic distortions and semantic preference which are usually insensitive to resolutions or frame rates, we follow the pre-processing in DOVER~\cite{dover} to perform \textbf{\textit{spatial down-sampling}} and \textbf{\textit{temporal sparse frame sampling}} on the original video. We denote the downsampled aesthetic-specific view of the video as $\mathcal{V} = \{V_i|_{i=0}^{N}\}$, where $V_i$ is the $i$-th frame (in total $N$ frames sampled) of the downsampled video, with spatial resolution $224\times224$, aligned with the spatial scale during the pre-training of CLIP~\cite{clip}. The spatial downsampling ensures the uncompromised understanding of semantic information in video frames, and shows more competitive performance than using full-resolution inputs~\cite{clipiqa} in various in-the-wild VQA datasets (compared in Tab.~\ref{tab:downsample}). %\footnote{The pre-processing is contrary to the design of Wang \textit{et al.}~\cite{clipiqa}, which attempts to retain the original resolution to keep low-level details about distortions. However, through experiments (Sec.~\ref{sec:downsample}), we prove that it is more \textbf{effective} and \textbf{efficient} to downsample videos in this branch.}

\subsubsection{Affinity between Video and Texts} Given any text prompt $T$, the visual ($\mathrm{E}_v$) and textual ($\mathrm{E}_t$) encoders in CLIP extract $\mathcal{V}$ and $T$ into global visual ($f_{v,i}^\mathrm{Global}$) and textual ($f_t$) features:
\begin{equation}
    f_{v,i}^\mathrm{Global} = \mathrm{E}_v(V_i)|_{i = 0}^{N-1}; ~~~~~~
    f_{t}^{T} = \mathrm{E}_t(T)
\end{equation}
Then, the semantic affinity $\mathrm{A}(\mathcal{V}, T)$ between $\mathcal{V}$ (the texture-insensitive view of the video) and text $T$ is defined by comparing the dot product between visual and text features:
\begin{equation}
    \mathrm{A}(\mathcal{V}, T) = (\sum_{i=0}^{N-1}\frac{f_{v,i}\cdot f_{t}^{T}}{\Vert f_{v,i}\Vert\Vert f_{t}^{T}\Vert})/N
    \label{eq:affinity}
\end{equation}
where the $\cdot$ denotes the dot product of two vectors.

\subsubsection{Antonym-Differential Affinity} In general, a video with good quality should be with higher affinity to \textbf{\green{positive}} quality-related descriptions or feelings ($T_+$, \textit{e.g. ``high quality", ``good", ``clear"}), and lower affinity to \bred{negative} quality-related text descriptions ($T_-$, \textit{e.g. ``low quality", "bad", "unclear"}, antonyms to $T_+$). Therefore, we introduce the Antonym-Differential affinity index ($\mathrm{DA}$), \textit{i.e.} whether the video has a higher affinity to positive or negative texts (Fig.~\ref{fig:criterion}(c)), as the semantic criterion for zero-shot VQA:
\begin{equation}
    \mathrm{DA}(\mathcal{V},T_{+},T_{-}) = \mathrm{A}(\mathcal{V}, T_{+}) -  \mathrm{A}(\mathcal{V}, T_{-})
\label{eq:ad}
\end{equation}
%where $T_+$ and $T_-$ are the positive and negative descriptions that form an antonym pair, \textit{e.g. good} and \textit{bad}.

\subsubsection{Selction of Prompts} Following the official recommendation of CLIP~\cite{clip} as well as several existing practices, we design the text prompts as a concatenation of a prefix, a description and a suffix. Specifically, the text prompt $T$ for raw description $D$ is defined as follows:
\begin{equation}
    T = \textit{`a '} + D + \textit{` photo'}
\end{equation}

The suffix is designed as \textit{``photo"} so as to drive the prompts to focus on visual quality while assigned with more general description pairs (\textit{good/bad}). Moreover, as we would like to extract both authentic distortions (which can hardly be detected by NIQE or other low-level indices) and aesthetic-related issues in the semantic quality index, we aggregate two different pairs of antonyms: \textbf{1)} (prone to distortion perception) \textit{a high quality photo}$\leftrightarrow$\textit{low quality photo}  ($T_{+}^{0}, T_{-}^{0}$); \textbf{2)} (prone to semantic goodness) \textit{a good photo}$\leftrightarrow$\textit{a bad photo} ($T_{+}^{1}, T_{-}^{1}$) into the multi-prompt differential affinity ($\mathrm{MPDA}$).  Finally, following the guidance of VQEG~\cite{vqeg} on perceptual scales of quality evaluation, we conduct sigmoid remapping to map the raw  $\mathrm{Q_{MPDA}}$ scores into range $[0,1]$, as the final SAQI ($\mathrm{Q}_{A}$):
\begin{align}
\mathrm{Q_{MPDA}} &= \sum_{d=0}^1 \mathrm{DA}(\mathcal{V},T_{+}^{d},T_{-}^{d}) \\
    \mathrm{Q}_{A} &= \frac{1}{1+e^{-\mathrm{Q_{MPDA}}}}
\label{eq:sigmoida}
\end{align}
%We evaluate effects of multi-prompt aggregation in Sec.~\ref{sec:ablprompt}.

\subsection{The Spatial Naturalness Index ($\mathrm{Q}_S$)}
\label{sec:qs} Despite the powerful SAQI, we also utilize the NIQE~\cite{niqe} index, the first completely-blind quality index to detect the traditional types of \textbf{technical distortions}, such as \textit{Additive White Gaussian Noises (AWGN), JPEG compression artifacts}. As distortions are very likely to happen in real-world videos, which suffer from bad compression or transmission qualities. It works by quantifying the difference between the input image features and the expected distribution of features for \textit{``high-quality''} summarized from various pristine natural images.

As raw NIQE scores ($\mathrm{Q}_{\mathrm{NIQE},i}$ for $V_i$) denote the ``raw" distance to the distribution of high quality videos, they are in a different scale range compared with the SAQI. To align the two indices, we \textit{normalize} them into Gaussian distribution $N(0,1)$ and rescale them with negative sigmoid-like remapping to get the frame-wise naturalness index ($\mathrm{N}_i$):

\begin{equation}
    \mathrm{N}_i = \frac{1}{1 + e^{\frac{\mathrm{Q}_{\mathrm{NIQE},i}-\overline{\mathrm{Q}_{\mathrm{NIQE},i}}}{\sigma(\mathrm{Q}_{\mathrm{NIQE},i})}}}
\label{eq:sigmoids}
\end{equation}
where $\overline{\mathrm{Q}_\mathrm{NIQE}}$ and $\sigma(\mathrm{Q}_\mathrm{NIQE})$ are the \textit{mean} and \textit{standard deviance} of raw NIQE scores in the whole set, respectively. Consequently, $\mathrm{N}_i$ also lies in range $[0,1]$. Then, following~\cite{tlvqm,videval,fastervqa}, we sample one frame per second (\textit{1fps}) and calculate the overall \textbf{Spatial Naturalness Index} ($\mathrm{Q}_S$) as follows:
\begin{equation}
    \mathrm{Q}_S = \sum_{k=0}^{S_0} \mathrm{N}_{F_k} / S_0
\end{equation}
where $S_0$ is the overall duration of the video, and $V_{F_k}$ is the ${F_k}$-th frame, sampled from the $k$-th second. %We do not conduct any spatial downsampling for $\mathrm{Q}_S$ to avoid corruption to quality-related information.

\subsection{The Temporal Naturalness Index ($\mathrm{Q}_T$)}
\label{sec:qt}

While the $\mathrm{Q}_A$ and $\mathrm{Q}_S$ can better cover different types of spatial quality issues, they are unable to cover the distortions in the temporal dimension, such as \textit{shaking}, \textit{stall}, or \textit{unsmooth camera movements}, which are well-recognized~\cite{cnn+lstm,tlvqm,deepvqa,discovqa} to affect the human quality perception. In general, all these temporal distortions can be summarized as non-smooth inter-frame changes between adjacent frames, and can be captured via recently-proposed TPQI~\cite{tpqi}, which is based on the neural-domain trajectory across three continuous frames. Specifically, the simulated neural responses on the primary visual cortex (V1,~\cite{primaryv1}) through the 2D Gabor filter~\cite{gabor} and lateral geniculate nucleus (LGN,~\cite{lgn}) domains for each frame is computed, and then the TPQI index is derived from curvatures from the two domains, formulated as follows:
\begin{equation}
    \mathrm{\mathrm{Q}_{TPQI}} = \frac{1}{2}\log{(\frac{1}{M-2}\sum_{j=1}^{M-2})\mathbb{C}^\mathrm{V1}_j} + \frac{1}{2}\log{(\frac{1}{M-2}\sum_{j=1}^{M-2})\mathbb{C}^\mathrm{LGN}_j}
\end{equation}
where $M$ is the total number of frames in the whole video, $\mathbb{C}_j^\mathrm{LGN}$ and $\mathbb{C}_j^\mathrm{V1}$ are the curvatures at a three-frame video-let $(j-1,j,j+1)$ respectively. The \textbf{Temporal Naturalness Index} ($\mathrm{Q}_T$) is then mapped from the raw scores via gaussian normalization and sigmoid rescaling:
\begin{equation}
    \mathrm{Q}_T = \frac{1}{1 + e^{\frac{\mathrm{Q}_{\mathrm{TPQI}}-\overline{\mathrm{Q}_{\mathrm{TPQI}}}}{\sigma(\mathrm{Q}_{\mathrm{TPQI}})}}}
\label{eq:sigmoidt}
\end{equation}
%To retain temporal distortions, we do not conduct any frame sampling here and feed all the frames to calculate $\mathrm{Q}_T$.

\begin{figure}
    \centering
\includegraphics[width=\linewidth]{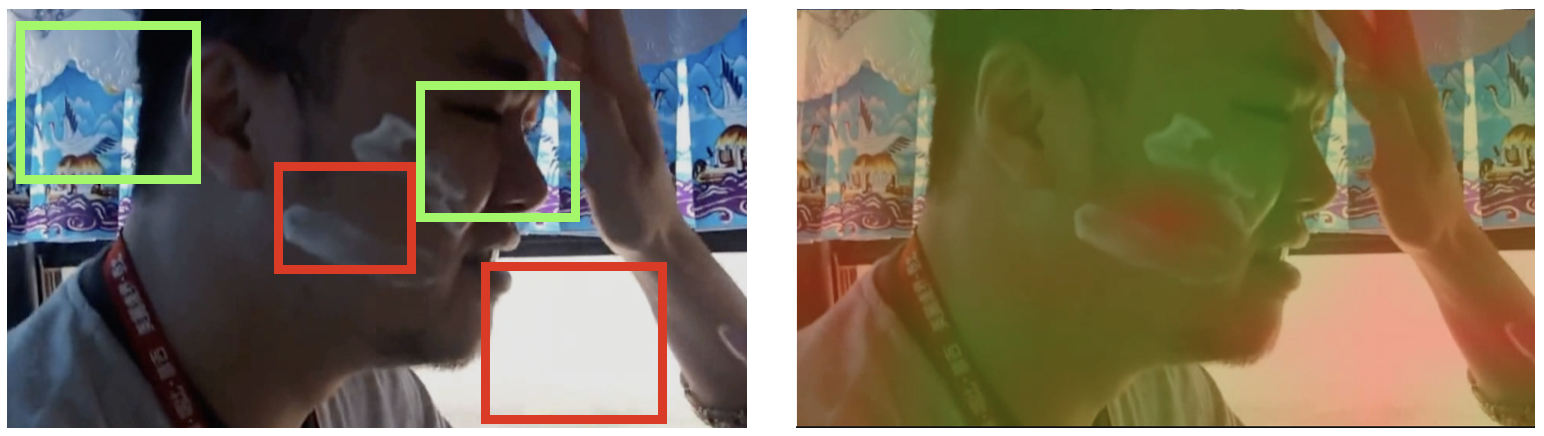}
    %\vspace{-9pt}
    \caption{\textbf{\textit{left}}: Spatial locality of quality information, where areas in \green{green} boxes have better quality than those in \red{red} boxes;  \textbf{\textit{right}}: visualization for \textbf{SAQI-Local} to predict localized semantic-related quality (more examples in Fig.~\ref{fig:lqm}).}
    \label{fig:locality}
    \vspace{-12pt}
\end{figure}

\subsection{\textit{BVQI} Index: Metric Aggregation}
As we aim to design a robust zero-shot perceptual quality index, we directly aggregate all the indices by summing up the scale-aligned scores without fine-tuning from any VQA datasets. As the $\mathrm{Q}_A$, $\mathrm{Q}_S$ and $\mathrm{Q}_T$ have already been gaussian-normalized and sigmoid-rescaled in Eq.~\ref{eq:sigmoida}, Eq.~\ref{eq:sigmoids} and Eq.~\ref{eq:sigmoidt} respectively, all three metrics are in range $[0,1]$, the overall unified \textbf{BVQI} index $\mathrm{Q}_\text{Unified}$ is defined as:
\begin{equation}
    \mathrm{Q}_\text{Unified} = \mathrm{Q}_A + \mathrm{Q}_S + \mathrm{Q}_T
    \label{eq:aggregation}
\end{equation}

\subsection{BVQI-Local}

\begin{figure*}
    \centering
    \includegraphics[width=0.93\textwidth]{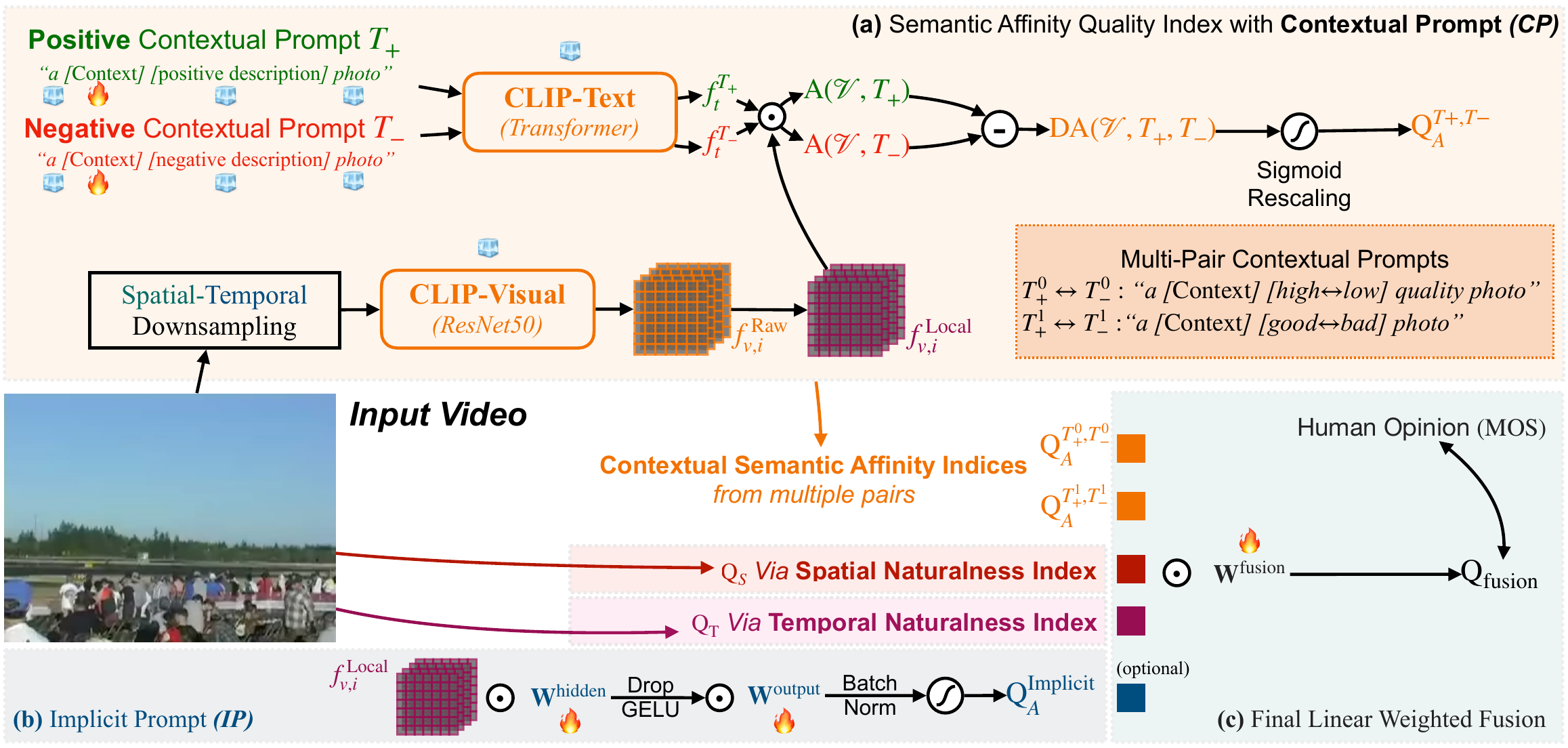} 
    %\vspace{-10bpt}
    \caption{The proposed approach for efficient dataset-specific fine-tuning based on \textbf{BVQI-Local}, including \textbf{(a)} Contextual Prompt (2K trainable parameters), \textbf{(b)} Implicit Prompt (65K trainable parameters), and \textbf{(c)} Final Linear Weighted Fusion (5 trainable parameters) of different indices.}
    \label{fig:finetune}
    \vspace{-10pt}
\end{figure*}

\subsubsection{Motivation: Locality of Quality Information} 

Several recent VQA studies~\cite{fastvqa,paq2piq,pvq} propose that quality information is localized, and that various regions of a video frame can have different quality levels, which collectively determine the global quality. During the construction of the semantic affinity criterion, we also observed that different spatial regions contain distinct semantic information (as illustrated in Fig.~\ref{fig:locality} \textbf{\textit{left}}) and exhibit varying distortion levels (\textit{e.g.} some regions are better-exposed than others). As a result, we aim to establish a localized semantic affinity criterion that evaluates the quality of different local regions, as discussed in further details below.

\subsubsection{Semantic Affinity Quality Localizer (SAQI-Local)}
\label{sec:local}
The default output of the visual encoder ($\mathrm{E}_v$) in CLIP computes the final visual features through an attention pooling layer $\mathrm{AttnPool}$. Given the raw features before the attention pooling ($f_{v,i}^\mathrm{Raw}$), the pooled local and global features are obtained through the multi-head self-attention~\cite{allyouneed} ($\mathrm{MHSA}$), as follows:
\begin{equation}
    {f_{v,i}^\mathrm{Global}}, f_{v,i}^\mathrm{Local} = \mathrm{MHSA}(\overline{f_{v,i}^\mathrm{Raw}}, f_{v,i}^\mathrm{Raw})\\
\end{equation}
where ${f_{v,i}^\mathrm{Global}}$ and $f_{v,i}^\mathrm{Local}$ are respective self-attention outputs from average-pooled features $\overline{f_{v,i}^\mathrm{Raw}}$ and raw features ${f_{v,i}^\mathrm{Raw}}$.

While only the ${f_{v,i}^\mathrm{Global}}$ is used during training of CLIP, the homogeneous characteristics of $\mathrm{MHSA}$ decide that the local features $f_{v,i}^\mathrm{Local}$ also contain valid semantic information. Therefore, similar as Eq.~\ref{eq:affinity}, we compute the local affinity ($\mathrm{LA}$) for each feature pixel, as follows:

\begin{equation}
    \mathrm{LA}(\mathcal{V}_{i,j,k}, T) = \frac{f_{v,i,j,k}^\mathrm{Local}\cdot f_{t}^{T}}{\Vert f_{v,i,j,k}^\mathrm{Local}\Vert\Vert f_{t}^{T}\Vert}
    \label{eq:laffinity}
\end{equation}
where $f_{v,i,j,k}^\mathrm{Local}$ means $i$-th local feature in row $j$, column $k$.

Then, the Semantic Affinity Quality Localizer (\textbf{SAQI-Local}, $\mathrm{Q}_{A,\text{Local}}$, visualized in Fig.~\ref{fig:locality} \textit{\textbf{right}}) is defined as:
\begin{equation}
    \mathrm{Q}_{A,\text{Local}_{i,j,k}} = \frac{1}{1+e^{-{\sum_{d=0}^1 \mathrm{LA}(\mathcal{V}_{i,j,k}, T_{+}^d) -  \mathrm{LA}(\mathcal{V}_{i,j,k}, T_{-}^d)}}}
\end{equation}
and $\mathrm{Q}_{A,\text{Local}}$ for all feature pixels in all frames are average pooled as the overall quality score for the video.

In addition to evaluating the overall perceptual quality of video regions, SAQI-Local has the capability to detect quality issues from various perspectives when more specific prompt pairs are defined (such as ``\textit{sharp/fuzzy}" or ``\textit{pleasant/annoying}", see Fig.~\ref{fig:lqm}). This targeted approach to localization results in more precise identification of quality issues.

\subsubsection{An Improved Overall Quality Index}

As the spatial and temporal naturalness indices require information for whole frames, it is unable to convert the two  into respective localized versions. Still, the proposed SAQI-Local can currently be integrated with the original versions of them for improved quality prediction of overall videos, by replacing the $\mathrm{Q}_A$ into $\mathrm{Q}_{A,\text{Local}}$ in Eq.~\ref{eq:aggregation}, denoted as \textbf{BVQI-Local}. Both the BVQI-Local and the SAQI-Local prove better alignment with human quality perception (see Tab.~\ref{tab:ablation}) than their global-feature-based counterparts, presenting that human quality perception is more likely to rely on collecting regional quality information.

\section{Efficient Dataset-Specific Fine-tuning}

The zero-shot index has achieved stable and excellent accuracy on various NR-VQA datasets. However, the definition of ``quality" in real-world scenarios can be ambiguous and may differ across different situations~\cite{dover}. Therefore, we have developed an efficient dataset-specific fine-tuning strategy. To better align the text prompts with specific scenarios, we propose the Contextual Prompt (Sec.~\ref{sec:cp}) to optimize the embedded text prompts, as well as the Implicit Prompt (Sec.~\ref{sec:ip}) to directly map the visual features to quality scores. Additionally, simply summing up separate indices ($\mathrm{Q}_A, \mathrm{Q}_S, \mathrm{Q}_T$) may not accurately reflect the tendencies or biases of different datasets. To address this issue, we introduce a Final Linear Weighted Fusion (Sec.~\ref{sec:lwf}) to aggregate different indices using a dataset-specific weighted sum. The fine-tuning pipeline requires fewer than 0.1M trainable parameters.

%Therefore, we propose an efficient low-rank fine-tuning pipeline of \textbf{BVQI-Local} to tackle with two challenges of the zero-shot index, as follows.

\subsection{Contextual Prompt} 
\label{sec:cp}

Due to the increasing scale of foundation models, it is becoming harder to fine-tune the whole models with limited computational resources. In recent years, prompt-tuning~\cite{softprompt,prefixprompt,coop,cocoop} strategies have been proposed, which keep the weights of the model \textbf{frozen} and only optimize the input text prompts. For the BVQI-Local, we follow \cite{coop} to design the learnable Contextual Prompt (\textbf{\textit{CP}}) $T_\mathrm{Ctx}$ for different VQA datasets:
\begin{equation}
    T_\mathrm{Ctx} = \textit{`a '} + \mathrm{[Context]} + D + \textit{` photo'}
\end{equation}
where the $[\mathrm{Context}]$ is designed as one single token, initialized as \textit{``X"} and optimized\footnote{As raw word tokens are discrete and cannot allow for back-propagation, in practice, we optimize the continuous embeddings of $\mathrm{[Context]}$. } during fine-tuning. Other parts of $T_\mathrm{Ctx}$ as well as \textbf{all weights in CLIP} are fixed to avoid over-fitting.

\subsection{Implicit Prompt}
\label{sec:ip}

%Several works~\cite{lpips,mlsp,svqa,nima} notice that perceptual quality opinions are hard to be totally explicitly reasoned. Therefore, we add a simple linear weighted layer ($\mathbf{W}^\mathrm{implicit}$) followed by sigmoid rescaling (as we have done for zero-shot indices) on the local visual features $f_{v,i}^\mathrm{Local}$ as the {Implicit Prompt} (\textbf{\textit{IP}}) for fine-tuning BVQI-Local, as follows:

%\begin{equation}
%\mathrm{Q}_A^\text{Implicit} = \frac{1}{1+e^{-{f_{v,i}^\mathrm{Local}^ \cdot \mathbf{W}^\mathrm{implicit}}}}
%\end{equation}
%where $\mathrm{Q}_A^\text{Implicit}$ is the output implicit-prompted quality score.

Several works~\cite{lpips,mlsp,svqa,nima} notice that perceptual quality opinions are hard to be totally explicitly reasoned. Therefore, we add an implicit multi-layer perception ($MLP$) followed by normalization and sigmoid rescaling (as we have done for zero-shot indices) on the local visual features $f_{v,i}^\mathrm{Local}$ as the {Implicit Prompt} (\textbf{\textit{IP}}), as follows:
\begin{equation}
\mathrm{Q}_A^\text{Implicit} = \frac{1}{1+e^{-{\mathrm{BatchNorm}(\mathrm{MLP}(f_{v,i}^\mathrm{Local}))}}}
\end{equation}
where $\mathrm{BatchNorm}$ is the batch normalization layer, and $\mathrm{Q}_A^\text{Implicit}$ is the output implicit-prompted quality score.

%Since $\mathrm{Q}_A^\text{Implicit}$ may compromise the interpretability of the final output, we evaluate two versions of the fine-tuned BVQI-Local: oThough former one has better intra-dataset performance, our analysis is conducted with the latter one (which has only 0.5\% less accuracy).

\subsection{Final Linear Weighted Fusion}
\label{sec:lwf}

Due to the differences in data distribution and biases of opinions under different situations, it is not always best to directly sum up all three indices for all VQA datasets. Moreover, we also would like different prompt pairs to be re-weighted based on different datasets. Therefore, we split the $\mathrm{Q}_{A, \text{Local}}$ into $\mathrm{Q}_A^{{T_+^0, T_-^0}}$ (for prompt pair \textit{[high $\leftrightarrow$low] quality}) \& $\mathrm{Q}_A^{{T_+^1, T_-^1}}$ (for pair \textit{[good $\leftrightarrow$bad]}) and design the dataset-specific final-linear weighted fusion as follows:
\begin{equation}
    \mathrm{Q}_\text{fusion} = [\mathrm{Q}_A^{{T_+^0, T_-^0}}, \mathrm{Q}_A^{{T_+^1, T_-^1}}, \mathrm{Q}_A^\text{Implicit}, \mathrm{Q}_S, \mathrm{Q}_T]^T \mathbf{W}^\mathrm{fusion}
\end{equation}
where $\mathbf{W}^\mathrm{fusion} \in \mathcal{R}^{5\times1}$ is the final fusion weight, jointed optimized with the contextual prompt and implicit prompt. For the variant without the implicit prompt, $\mathbf{W}^\mathrm{fusion}\in \mathcal{R}^{4\times1}$.

 %the two naturalness indices are naturally focusing on information in a small receptive field~\cite{niqe,tpqi} and can be easily adapted to the localized version through patch-wise inputs. 

\section{Experimental Evaluations}

In this section, we mainly answer several important questions about the proposed zero-shot quality indices as well as the dataset-specific efficient fine-tuning process.

\begin{itemize}
    \item Is the proposed method efficient enough, in terms of both zero-shot inference and fine-tuning cost (Sec~\ref{sec:efficiency})?
    \item What is the accuracy of the proposed zero-shot (\textit{w/o fine-tuning}) quality indices (Sec~\ref{sec:benchmark})?
    \item After fine-tuning, can the BVQI-Local outperform existing methods while retaining high robustness (Sec.~\ref{sec:finetunes})?
    \item Analysis (Sec.~\ref{sec:analysis}): What are the quality concerns of SAQI? How do they differ from traditional metrics?
    \item What are the effects (Sec.~\ref{sec:ablation}) of spatial-temporal downsampling in SAQI, separate indices, prompt design, and designs in the proposed fine-tuning scheme (Sec.~\ref{sec:ablft})?
\end{itemize}

\begin{table*}
\footnotesize
\caption{Benchmark between the proposed zero-shot BVQI (BVQI-Local) and existing zero-shot quality indices. The fine-tuned BVQI-Local is further compared with existing training-based methods. For fairness, methods~\cite{bvqa2021,fastvqa} that include extra IQA/VQA annotated data for training are excluded.}\label{table:eva}
\label{table:vqc}
\setlength\tabcolsep{9.8pt}
\renewcommand\arraystretch{1.2}
\footnotesize
\centering
%\vspace{-8pt}
\resizebox{\textwidth}{!}{\begin{tabular}{l|cc|cc|cc|cc}
\hline
\textbf{Dataset}     & \multicolumn{2}{c|}{LIVE-VQC}   & \multicolumn{2}{c|}{KoNViD-1k}        & \multicolumn{2}{c|}{YouTube-UGC}     &  \multicolumn{2}{c}{CVD2014}           \\ \hline
Methods
        & SRCC$\uparrow$   & PLCC$\uparrow$      & SRCC$\uparrow$   & PLCC$\uparrow$             & SRCC$\uparrow$   & PLCC$\uparrow$                   &SRCC$\uparrow$  & PLCC$\uparrow$                                 \\ \hline 
%\rowcolor{lightgray} \multicolumn{9}{l}{\textbf{Opinion-Aware Methods, Extra Data Used:} }           \\ \hdashline
%\rowcolor{lightgray} Li \textit{et al.} (TCSVT, 2022) \cite{bvqa2021}  & 0.834 & 0.842 & 0.834 & 0.836 & 0.818 & 0.826 & 0.858 & 0.873  \\ 
%\rowcolor{lightgray} {DOVER} (Arxiv, 2022) \cite{dover}  &  {{0.858}} & {{0.874}} & {{0.906}} & {{0.905}} & {{0.880}} & {{0.874}}  & {{0.894}} & {{0.908}} \\ \hdashline
\multicolumn{9}{l}{\textbf{(a) Zero-shot Quality Indices:} }           \\\hdashline

(\textit{Spatial}) NIQE (Signal Processing, 2013)~\cite{niqe}  & 0.596 & 0.628 & 0.541 & {0.553} & 0.278 & 0.290 & {0.492} & {0.612} \\
(\textit{Spatial}) IL-NIQE (TIP, 2015)~\cite{ilniqe} & 0.504 & 0.544 & 0.526 & 0.540  & {0.292} & {0.330} & 0.468 & 0.571\\
(\textit{Temporal}) VIIDEO (TIP, 2016)~\cite{viideo} & 0.033 & 0.215 & 0.299 & 0.300  & 0.058 & 0.154 & 0.149 & 0.119 \\
(\textit{Temporal}) TPQI (ACMMM, 2022)~\cite{tpqi}  & {0.636} & {0.645} & \textbf{0.556} & 0.549 & 0.111 & 0.218 & 0.408 & 0.469 \\ 
\hdashline
%{naïve CLIP-based Quality Index}* (baseline) &0.608&0.581&0.586&0.551&0.473&0.458&0.507&0.512\\
{(\textit{Semantic}) SAQI (\textit{Ours, ICME2023})} & 0.629 & 0.638 & 0.608 & 0.602 & \blue{0.585} & \blue{0.606} & 0.685 & 0.692 \\
{(\textit{Semantic}) SAQI-Local (\textit{Ours, extended})} & \textbf{0.651} & \textbf{0.663} & \textbf{0.622} & \textbf{0.620} & \bred{0.610} & \bred{0.616} & \textbf{0.734} & \textbf{0.731} \\\hdashline
\rowcolor{lightpink} (\textit{Aggregated}) \textbf{BVQI} (\textit{Ours, ICME2023}) & \blue{0.784} & \blue{0.794} & \blue{0.760} & \blue{0.760} & {0.525} & {0.556} & \blue{0.740} & \blue{0.763}\\ 

\rowcolor{lightpink} (\textit{Aggregated}) \textbf{BVQI-Local} (\textit{Ours, extended}) & \bred{0.794} & \bred{0.803} & \bred{0.772} & \bred{0.772} & \textbf{0.550} & \textbf{0.563} & \bred{0.747} & \bred{0.768}\\  \hline
 \multicolumn{9}{l}{\textbf{(b) Fine-tuned VQA Methods:} }           \\ \hdashline
% \multicolumn{7}{|c|}{\textit{Existing Methods}}  \\ \hline 
 TLVQM (TIP, 2019) \cite{tlvqm}    & 0.799 &  0.803  & 0.773 & 0.768   & 0.669 &  0.659   & 0.830 &    0.850     \\
 VSFA (ACMMM, 2019) \cite{vsfa}         & 0.773 &  0.795  & 0.773 & 0.775   & 0.724 &  0.743   & 0.870 &     0.868 \\
 CNN-TLVQM (ACMMM, 2020) \cite{videval} &  0.825 & 0.834 & 0.816 & 0.818 & 0.809 & 0.802 & 0.857 &  0.869 \\ 
 VIDEVAL (TIP, 2021) \cite{videval} &  0.752 &  0.751  & 0.783 & 0.780           & 0.779 &  0.773    & 0.832 &     0.854   \\ 
PVQ (CVPR, 2021) \cite{pvq} &  0.827 &  0.837  & 0.793 & 0.705           & 0.790 &  0.791    & 0.866 &     0.874   \\
CoINVQ (CVPR, 2021) \cite{rfugc} & NA & NA & 0.767 & 0.762 & \bred{0.816} & {0.802} & NA & NA \\
GST-VQA (TCSVT, 2022) \cite{gstvqa} & 0.801 & 0.805 & 0.814 & 0.825 & 0.797 & 0.792 & 0.831 & 0.844 \\
\hdashline
\rowcolor{lightcyan} \textbf{BVQI-Local} + \textbf{\textit{CP}} (\underline{C}ontextual \underline{P}rompt) & \blue{0.832} & \blue{0.844} & \blue{0.827} & \blue{0.831} & \blue{0.808} & \blue{0.803} & \blue{0.871} & \blue{0.877} \\ 
\rowcolor{lightcyan} \textbf{BVQI-Local} + \textbf{\textit{CP}} + \textbf{\textit{IP}} (\underline{I}mplicit \underline{P}rompt) & \bred{0.840} & \bred{0.850} & \bred{0.833} & \bred{0.834} & \bred{0.816} & \bred{0.804} & \bred{0.876} & \bred{0.882} \\ \hline
%\textit{In-distribution Gain of Fine-tuning} & +0.046 & +0.047 & +0.061 & +0.062 & +0.266 & +0.241 & +0.130 & +0.115\\ \hline
\end{tabular}}
\vspace{-13pt}
\end{table*}

\subsection{Evaluation Settings}

\subsubsection{Implementation Details} Due to the differences in the targeted quality-related issues in the three indices, the inputs of the three branches are different. For $\mathrm{Q}_A$, the video is spatially downsampled to $224\times 224$ via a bicubic~\cite{bicubic} downsampling kernel, and temporally sub-sampled to $N=32$ uniform frames~\cite{dover}. For $\mathrm{Q}_S$, the video retains its original spatial resolution but temporally only keeps $S_0$ uniform frames, where $S_0$ is the duration of the video (\textit{unit: second}). For $\mathrm{Q}_T$, all videos are spatially downsampled to short-size $270$ and kept with the original aspect ratio, with all frames fed into the neural response simulator. The $\mathrm{Q}_A$ is calculated with Python 3.10, Pytorch 1.13, with official CLIP-ResNet-50~\cite{he2016residual} weights. The $\mathrm{Q}_S$ and $\mathrm{Q}_T$ are calculated with Matlab R2022b, while we also provided an equivalent Pytorch accelerated version for the two indices. The machine is with two E5 2678-v3 CPUs, one Tesla P40 GPU, with 64GB Memory and 24GB Graphic Memory. During fine-tuning, the batch size is set as 16, with 10 random 8:2 train-test splits divided by random seeds $\{i\times 42|_{i=1}^{10}\}$.

\begin{table}[]
\setlength\tabcolsep{8.5pt}
\renewcommand\arraystretch{1.2}
\caption{Inference FLOPs and time consumption for one 8-sec, 540P video. The speed difference between BVQI and BVQI-Local is negligible.} \label{tab:latency}
\vspace{-7pt}
\center
\footnotesize
\resizebox{\linewidth}{!}{\begin{tabular}{l|c|c}
\hline
Quality Index & GPU-Time(\textit{sec}) & CPU-Time(\textit{sec}) \\ \hline
Semantic Affinity (SAQI, $\mathrm{Q}_A$) & 0.264 & 5.78 \\ 
Spatial Naturalness ($\mathrm{Q}_S$) & 0.051 & 1.84 \\ 
Temporal Naturalness ($\mathrm{Q}_T$) &  0.685 & 47.31\\ \hdashline
\textbf{BVQI} (overall, $\mathrm{Q}_\text{Unified}$)  & 0.902 & 54.93 \\ \hline
\textit{- time consumption per frame} &  0.004 (\textit{266fps}) & 0.275 (\textit{3.63fps}) \\ \hline
\end{tabular}}
\vspace{-7pt}
\end{table}

\begin{table}[]
\setlength\tabcolsep{7.6pt}
\renewcommand\arraystretch{1.2}
\caption{Trainable parameters in two versions of fine-tuned BVQI-Local, compared with the frozen parameters in different parts of CLIP.} \label{tab:params}
\vspace{-7pt}
\center
\footnotesize
\resizebox{\linewidth}{!}{\begin{tabular}{l|c|c}
\hline
Module & \#Parameters & Relative Percentage \\ \hline
\multicolumn{3}{l}{\textbf{Frozen Parameters in CLIP}~\cite{clip}:}  \\ \hdashline
(\textit{CLIP-Text}) Token Embedding & 25,296,896 & 24.93\% \\
(\textit{CLIP-Text}) Transformer & 37,828,608 & 37.29\% \\
(\textit{CLIP-Visual}) Modified-ResNet-50 & 38,316,896
 & 37.77\% \\ \hdashline
\multicolumn{3}{l}{\textbf{Trainable Parameters in during efficient fine-tuning}:}  \\ \hdashline
Contextual Prompt (\textbf{\textit{CP}}) & 2,048 & 0.002\% \\
Final Linear Weighted Fusion & 4$_{\textit{w/o \textbf{IP}}}$/5 $_{\textit{w/ \textbf{IP}}}$ &  0.000\% \\
Total for \textbf{BVQI-Local + \textit{CP}} & \textbf{2,052} & \textbf{0.002\%} \\ \hdashline
Implicit Prompt (\textbf{\textit{IP}}) & 65,600 & 0.065\% \\
Total for \textbf{BVQI-Local + \textit{CP} + \textit{IP}} & \textbf{67,653} & \textbf{0.073\%} \\
\hline
\end{tabular}}
\vspace{-7pt}
\end{table}

\subsubsection{Evaluation Metrics} Following common studies, we use two metrics, the Spearman Rank-order Correlation Coefficients (SRCC) to evaluate monotonicity between quality scores and human opinions, and the Pearson Linearity Correlation Coefficients (PLCC) to evaluate linear accuracy. 
%Given predictions as $\mathrm{Q}_\mathrm{pred}$ and human mean opinion scores as $\mathrm{MOS}$, the two evaluation metrics are calculated as follows:
%\begin{align}
%    \mathrm{PLCC} &=  \frac{(\mathrm{Q}_\mathrm{pred} - \overline{\mathrm{Q}_\mathrm{pred}}) \cdot  (\mathrm{MOS} - \overline{\mathrm{MOS}})}{\Vert \mathrm{Q}_\mathrm{pred} - \overline{\mathrm{Q}_\mathrm{pred}} \Vert_2\Vert \mathrm{MOS} - \overline{\mathrm{MOS}} \Vert_2}) \\
%    \mathrm{SRCC} &= 1 - \frac{6\sum_{i=1}^N d_i^2}{N(N^2 - 1)}, 
%\end{align}

\subsubsection{Benchmark Datasets} To better evaluate the performance of the proposed BVQI and BVQI-Local under different in-the-wild settings, we choose four different datasets, including \textbf{CVD2014}~\cite{cvd} (234 videos, with lab-collected authentic distortions during capturing), \textbf{LIVE-VQC}~\cite{vqc} (585 videos, recorded by smartphones), \textbf{KoNViD-1k}~\cite{kv1k} (1200 videos, collected from social media platforms), and \textbf{YouTube-UGC}~\cite{ytugc,ytugccc} (1147 available videos, containing non-natural videos collected from YouTube with categories \textit{Screen Contents/Gaming/Animation/Lyric Videos}). 

\subsection{Efficiency}
\label{sec:efficiency}

\subsubsection{Inference Speed} In Tab.~\ref{tab:latency}, we show that the proposed BVQI index has very high inference speed. First, for its deep branch (the SAQI), the video is spatially and temporally downsampled, thus inference time is compressed to only 0.264 second on GPU (including the data pre-processing time). The main performance bottleneck comes from the temporal naturalness index (TPQI), where the Gabor filter requires weakly-paralleled computations on the complext domain. Still, the whole index requires less than one second to infer a 540P, 8-sec video on GPU, which is \textit{266fps} and 9 times faster than the standard of real-time inference.

\subsubsection{Training Parameters, Memory Cost and Speed} Compared with the original CLIP model which has over 100M parameters, the proposed efficient fine-tuning only needs to optimize 2K (without implicit prompt) or 67K (with implicit prompt), less than 0.1\% of total parameters of CLIP. Moreover, since all the backbone weights are fixed, the visual features and text embeddings can be pre-extracted and stored, further reducing the computational load during training. On our device, the fine-tuning only requires only \textbf{2.1GB Graphic Memory} cost with batch size 16, and need less than \textbf{2 minutes} to finish 30 epochs of tuning on KoNViD-1k (1,200 videos) dataset.%\footnote{The feature extraction on KoNViD-1k~\cite{kv1k} needs around 18 minutes and the extracted features can be re-used for multiple runs.}.

\begin{table}[]
\setlength\tabcolsep{5pt}
\renewcommand\arraystretch{1.26}
\caption{Fine-tuning results on the LSVQ~\cite{pvq} dataset. Though with very few parameters, the fine-tuning scheme can perform well on large datasets.} \label{tab:lsvq}
\vspace{-7pt}
\center
\footnotesize
\resizebox{\linewidth}{!}{\begin{tabular}{l|cc|cc}
\hline
\textbf{Train} on & \multicolumn{4}{c}{{LSVQ$_\text{train}$}} \\ \hline
\textbf{Test} on & \multicolumn{2}{c|}{{LSVQ$_\text{Test}$}} & \multicolumn{2}{c}{{{LSVQ$_\text{1080P}$}}}  \\ \hline
\textit{}                         & SRCC$\uparrow$                        & PLCC$\uparrow$                        & SRCC$\uparrow$                        & PLCC$\uparrow$  \\  \hline
BRISQUE (2013, TIP)~\cite{brisque} &  0.579 & 0.576 & 0.497 & 0.531       \\ 
TLVQM (2019, TIP)~\cite{tlvqm} & 0.772 & 0.774 & 0.589 & 0.616 \\
VIDEVAL (2021, TIP)~\cite{videval} & 0.794 & 0.793 & 0.545 & 0.554 \\
PVQ (2021, CVPR)~\cite{pvq} & 0.814 & 0.816 & 0.686 & 0.708\\
PVQ$_\textit{with extra patch labels}$ & 0.827 & 0.828 & 0.711 & 0.739 \\ \hline
\textbf{BVQI-Local} + \textbf{\textit{CP}}  \textit{(ours)} & \blue{0.838} & \blue{0.838} & \blue{0.738} & \blue{0.776} \\ 

\textbf{BVQI-Local} + \textbf{\textit{CP}} + \textbf{\textit{IP}} \textit{(ours)} & \bred{0.843} & \bred{0.843} & \bred{0.742} & \bred{0.782} \\ \hline

\end{tabular}}
\vspace{-7pt}
\end{table}

\begin{table*}[]
\setlength\tabcolsep{4.5pt}
\renewcommand\arraystretch{1.26}
\caption{Cross-dataset generalization evaluation. Even after fine-tuning on one dataset, the BVQI-Local can typically retain high accuracy on other datasets (compared with the zero-shot version), and show much better cross-dataset performance than existing approaches.} \label{tab:crossvsbv}
%\vspace{-17pt}
\center
\footnotesize
\resizebox{\linewidth}{!}{\begin{tabular}{l|cc|cc|cc|cc|cc|cc}
\hline
\textbf{Train} on & \multicolumn{4}{c|}{{KoNViD-1k}} & \multicolumn{4}{c|}{{LIVE-VQC}} & \multicolumn{4}{c}{{Youtube-UGC}} \\ \hline
\textbf{Test} on & \multicolumn{2}{c|}{{LIVE-VQC}} & \multicolumn{2}{c|}{{Youtube-UGC}}  & \multicolumn{2}{c|}{{KoNViD-1k}} & \multicolumn{2}{c|}{{Youtube-UGC}}   & \multicolumn{2}{c|}{{LIVE-VQC}} & \multicolumn{2}{c}{{KoNViD-1k}} \\ \hline
\textit{}                         & SRCC$\uparrow$                        & PLCC$\uparrow$                        & SRCC$\uparrow$                        & PLCC$\uparrow$                 & SRCC$\uparrow$                        & PLCC$\uparrow$                        & SRCC$\uparrow$                        & PLCC$\uparrow$        & SRCC$\uparrow$                        & PLCC$\uparrow$               & SRCC$\uparrow$                       & PLCC$\uparrow$                                         \\ 
\hline
TLVQM (2019, TIP)\cite{tlvqm} & 0.573 & 0.629 & 0.354 & 0.378 & 0.640  & 0.630& 0.218 & 0.250 & 0.488 & 0.546 & 0.556 & 0.578 \\
CNN-TLVQM (2020, MM)\cite{cnntlvqm}             & 0.713                      & 0.752    & \textbf{0.424} &   \textbf{0.469}               & 0.642 & 0.631 & 0.329 & 0.367      & 0.551 & 0.578 & 0.588 & 0.619                               \\ 

VIDEVAL (2021, TIP)\cite{videval}                    & 0.627                         & 0.654   & 0.370 & 0.390                  & 0.625                        & 0.621 & 0.302 & 0.318  & 0.542                         & 0.553      & 0.610 &   0.620                           \\ 
MDTVSFA (2021, IJCV)\cite{mdtvsfa}                     & \textbf{0.716}                        & \textbf{0.759}    & {0.408} &  {0.443}                 & 0.706                         & \textbf{0.711}  & \textbf{0.355} &  \textbf{0.388}      & \textbf{0.582}                        & \textbf{0.603}   & \textbf{0.649} &   \textbf{0.646}                     \\ 
GST-VQA (2022, TCSVT)\cite{gstvqa}             & 0.700                      & 0.733    & \gray{NA} &   \gray{NA}               & \textbf{0.709} & 0.707 & \gray{NA} & \gray{NA}       & \gray{NA}                        & \gray{NA}             & \gray{NA}  &   \gray{NA}                                 \\ \hdashline

\rowcolor{lightgray} \textbf{BVQI-Local} (\textit{before fine-tuning}) & \gray{0.794}&\gray{0.803}&\gray{0.550}&\gray{0.563}&\gray{0.772}&\gray{0.772}&\gray{0.550}&\gray{0.563}&\gray{0.794}&\gray{0.803}&\gray{0.772}&\gray{0.772}\\ \hdashline
\textbf{BVQI-Local} + \textbf{\textit{CP}} & \blue{0.776} & \bred{0.806} &\bred{0.653}&\bred{0.681}& \bred{0.778}&\bred{0.780}
&\bred{0.522} & \bred{0.552} &\blue{0.734}&\blue{0.751}&\blue{0.770}&\blue{0.767} \\ 
%\textit{Out-of-distribution Gain(+)/Loss(-)} & \textit{-0.018} & \textit{+0.003} & \textit{+0.103} & \textit{+0.118} & \textit{+0.006} & \textit{+0.008} & \textit{-0.028} & \textit{-0.011} & \textit{-0.060} & \textit{-0.050} & \textit{-0.005} & \textit{-0.008} \\ 
%\hdashline
\textbf{BVQI-Local} + \textbf{\textit{CP}} + \textbf{\textit{IP}} & \bred{0.782} & \bred{0.806} &\blue{0.650}&\blue{0.671}& \blue{0.770}&\blue{0.772}
&\blue{0.488} & \blue{0.515} &\bred{0.749}&\bred{0.764}&\bred{0.787}&\bred{0.787} \\ 
%\textit{Out-of-distribution Gain(+)/Loss(-)} & \textit{-0.012} & \textit{+0.003} & \textit{+0.100} & \textit{+0.108} & \textit{-0.002} & \textit{+0.000} & \textit{-0.062} & \textit{-0.048} & \textit{-0.045} & \textit{-0.039} & \textit{+0.015} & \textit{+0.015}  \\ 
\hline
\end{tabular}}
%\vspace{-7pt}
\end{table*}

\subsection{Zero-Shot Evaluation}
\label{sec:benchmark}

To evaluate the performance of the proposed BVQI and BVQI-Local, we evaluate it without fine-tuning in Tab.~\ref{table:eva}(a), in comparison of representative existing zero-shot VQA methods. The proposed BVQI is notably better than any existing zero-shot quality indices with \textbf{\textit{at least 20\%}} improvements on any dataset, while BVQI-Local steadily further improves the performance. It is also noteworthy that the proposed SAQI (before consideration of temporal quality and spatial details) can alone outperform all existing indices. The overall index can even be on par with or better than some fine-tuned approaches on the three natural VQA datasets (LIVE-VQC, KoNViD-1k and CVD2014). On the non-natural dataset (YouTube-UGC), with the assistance of powerful SAQI, the proposed BVQI-Local has extraordinary \textbf{88\%} improvement than all semantic-unaware zero-shot quality indices, \textit{for the first time} provides reasonable quality predictions on this dataset. Without fitting to any of the datasets, these results demonstrate that the proposed method achieves leapfrog improvements over existing metrics and can be widely applied as a robust real-world video quality metric.

\subsection{Evaluation on Fine-tuned Versions}
\label{sec:finetunes}

In this part, we evaluate the two fine-tuned versions of BVQI-Local, including the version which keeps the structure of original BVQI-Local (without implicit prompt, \textbf{+\textit{CP}}), and the full version with the implicit prompt (denoted as \textbf{+\textit{CP}+\textit{IP}}). 

\subsubsection{\textbf{Intra-dataset Evaluation}} After fine-tuning, both versions of BVQI-Local reach state-of-the-art or comparable performance on all VQA datasets, where the \textbf{\textit{+IP}} version performs slightly better. Moreover, we notice that fine-tuned versions has an average of 16\% of intra-dataset (in-distribution) performance gain than the zero-shot versions, proving that undoubted effectiveness of dataset-specific fine-tuning. As the fine-tuned BVQI-Local is with almost identical inference speed (as in Tab.~\ref{tab:latency}) to the zero-shot version, it is also more efficient than all other listed methods. This further proves the practical value of the fine-tuning the proposed quality indices. We also take a look at the results on a recently-proposed larger-scale dataset, LSVQ~\cite{pvq} (with 39,075 videos) in Tab.~\ref{tab:lsvq}, where the proposed lightweight fine-tuning can also reach competitive performance (though it is designed for small datasets), proving its potential scalability.

\subsubsection{\textbf{Cross-dataset Evaluation}} In Tab.~\ref{tab:crossvsbv}, we evaluate the cross-dataset generalization ability of different opinion-driven VQA methods. From the table, we reach three important observations: \textbf{1)} the zero-shot BVQI/BVQI-Local can already be provide better prediction on any dataset than existing methods trained on other datasets; \textbf{2)} while reaching much better alignment into one dataset, the proposed efficient fine-tuning will still retain to be effectively aligned with other datasets, and in average the fine-tuned BVQI-Local performs even slightly better than its zero-shot counterpart; \textbf{3)} henceforth, both two versions of fine-tuned BVQI-Local have extraordinary cross-dataset generalization ability, far more robust (\textit{+20\% improvement in average}) than existing methods.

\begin{figure}
    \centering
\includegraphics[width=\linewidth]{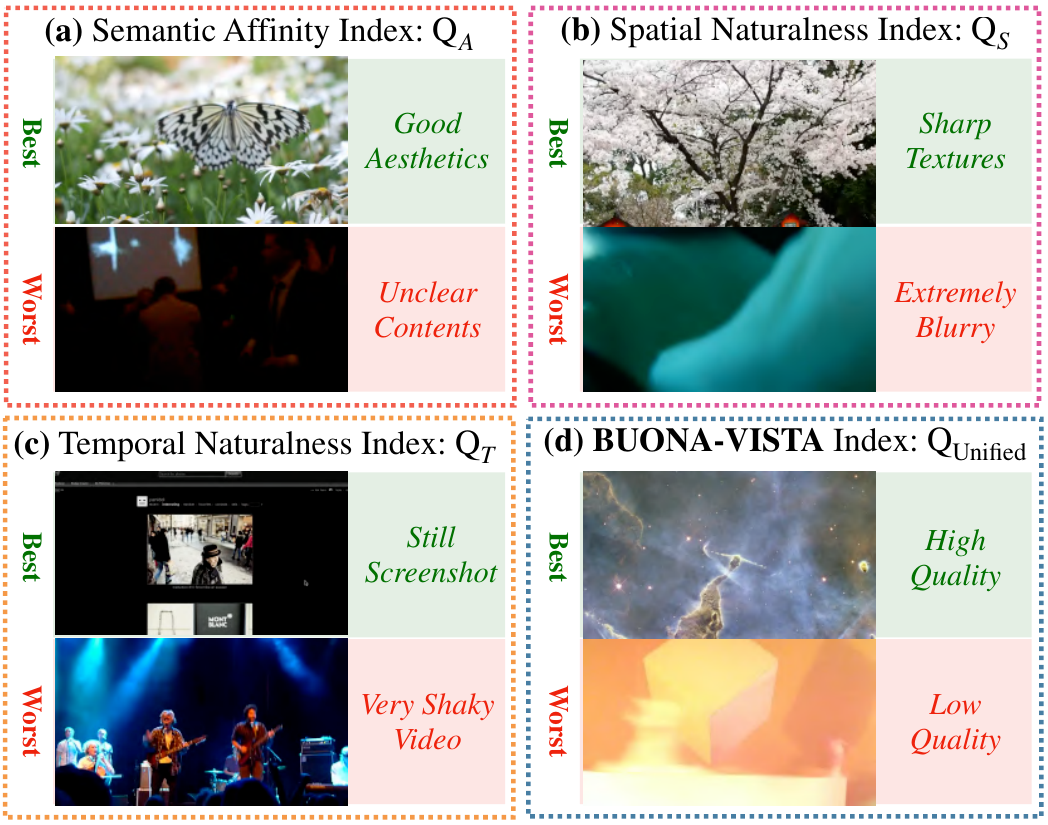}
    %\vspace{-18pt}
    \caption{Videos with \textit{best/worst} quality in perspective of three separate indices, and the overall BVQI (BUONA-VISTA). All demo videos are in our website.}
    \label{fig:vis}
    \vspace{-8pt}
\end{figure}

\begin{figure*}
 \includegraphics[width=\textwidth]{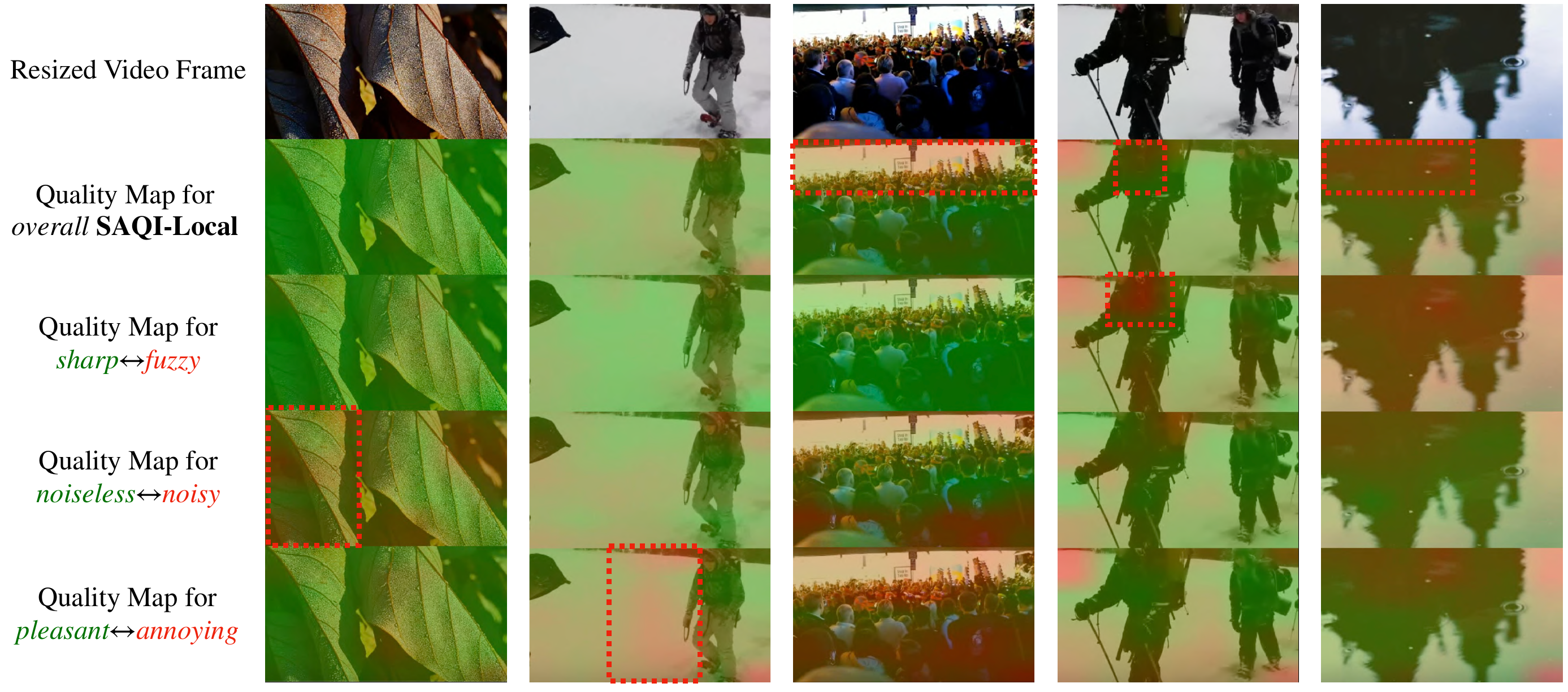}
    \caption{Localized quality maps in SAQI-Local (Sec.~\ref{sec:local}) for KoNViD-1k~\cite{kv1k}, where \green{green} areas refer to better quality, \red{red} areas refer to worse quality, and the area with worst quality are bounded in \red{red} dashed boxes. Results from both the default SAQI-Local and concrete prompt pairs are shown.}
    \label{fig:lqm}
    \vspace{-6pt}
\end{figure*}

\begin{figure}
    \centering
\includegraphics[width=0.98\linewidth]{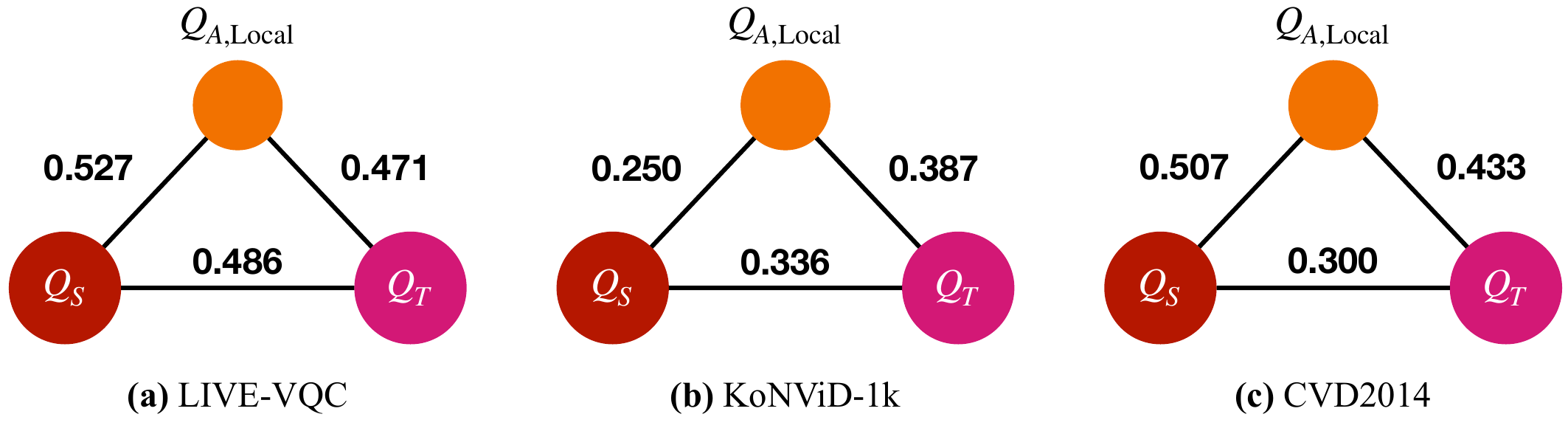}
    %\vspace{-18pt}
    \caption{\textbf{PLCC} (linear correlation) among three indices in LIVE-VQC, KoNViD-1k, and CVD2014 datasets are low, suggesting their divergence.}
    \label{fig:cor}
    \vspace{-12pt}
\end{figure}

\subsection{Analysis}
\label{sec:analysis}

\subsubsection{Best and Worst Videos in Each Index} In the first part of analysis, we visualize snapshots of videos with highest or lowest score in each separate index, and the overall BVQI, from the KoNViD-1k dataset. As shown in Fig.~\ref{fig:vis}, the \textbf{(a)} Semantic Affinity is highly related to \textbf{\textit{aesthetics}} (content appealingness), where the \textbf{(b)} Spatial Naturalness focus on spatial textures (\textit{sharp}$\leftrightarrow$\textit{blurry}), and the \textbf{(c)} Temporal Naturalness focus on temporal variations (\textit{stable}$\leftrightarrow$\textit{shaky}), aligning with the aforementioned criteria of the three indices. We also append the original videos of the examples in our website.

\subsubsection{Correlations Among Indices} Another evidence that the three indices are focusing on different parts of video quality is to examine the correlations among these indices, as illustrated in Fig.~\ref{fig:cor}. In general, these cross-index correlations are less than 0.5 PLCC, indicating that they are not so correlated with one another. The correlation values are also less than their correlation to the ground truth $\mathrm{MOS}$ (see Tab.~\ref{tab:ablation}), suggesting that they assess video quality from different perspectives.

\subsubsection{Localized Quality Maps} 
\begin{figure*}
 \includegraphics[width=\textwidth]{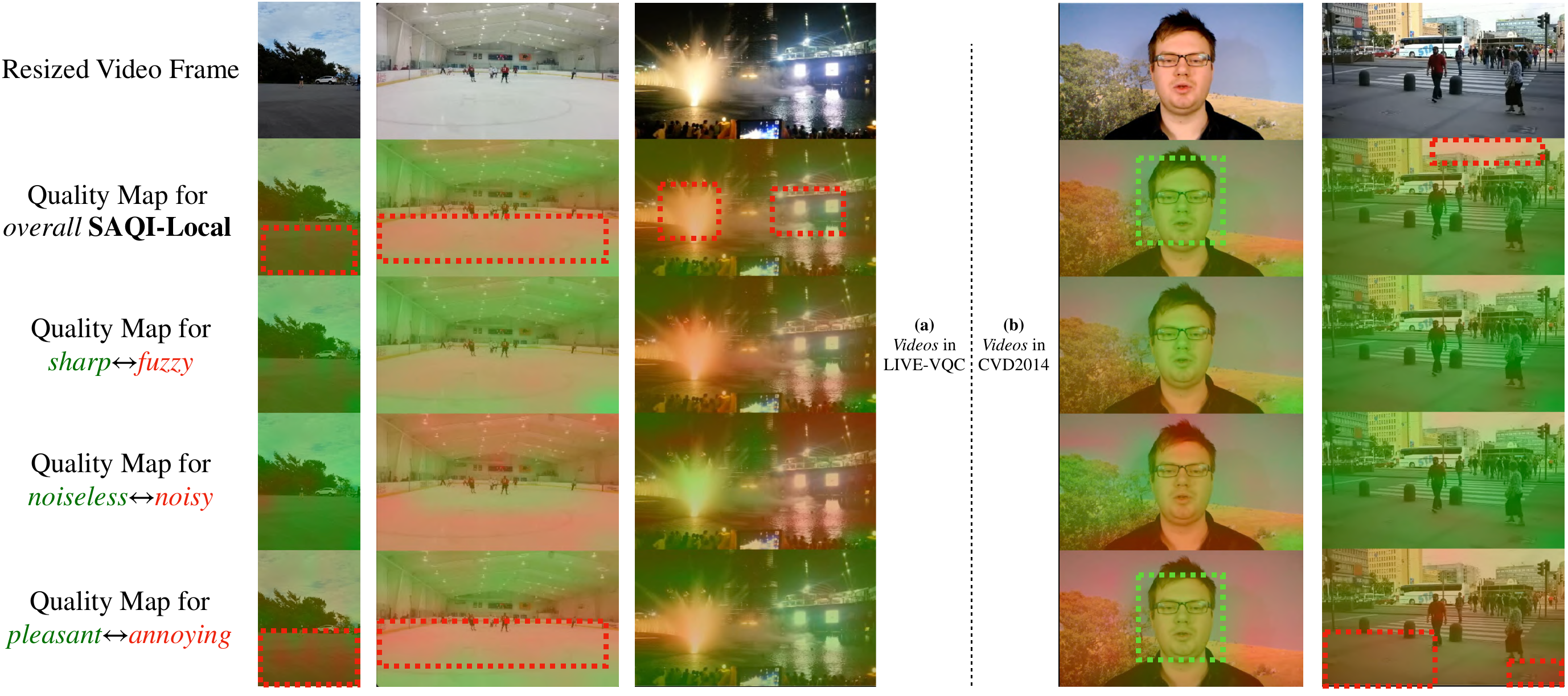}
    \caption{More localized quality maps (\textit{with the same legend as Fig.~\ref{fig:lqm}, \green{green} for better, \red{red} for worse}) for \textbf{(a)} LIVE-VQC~\cite{vqc} and \textbf{(b)} CVD2014~\cite{cvd}. These examples straightly show that the SAQI can distinguish between \textit{white clouds} and \textit{over-exposed areas}. }
    \label{fig:morelqm}
    \vspace{-6pt}
\end{figure*}

In Fig.~\ref{fig:lqm}, we show several examples of localized quality maps of SAQI-Local, where each video is presented with its original appearance as well as derived quality maps from the full SAQI-Local index and several concrete single prompt pairs, including \textit{[pleasant$\leftrightarrow$annoying]}, \textit{[sharp$\leftrightarrow$fuzzy]}, and \textit{[noise-free$\leftrightarrow$noisy]} (see their quantitative results in Tab.~\ref{tab:extprompt}). For the leftmost one, the video is in general with very good quality (good sharpness, clear and pleasant contents), yet there exists several noises, which could be detected for the prompt pair \textit{[noise-free$\leftrightarrow$noisy]}. For the three in the middle, we can notice that \textit{over/under exposure} and \textit{lack of meaningful contents} can both be well-captured by SAQI-Local. More importantly, it can distinguish between the dull background (\textit{snow}) and over-exposed areas, proving its strong semantic perception ability. In the rightmost video of a water reflection, SAQI-Local is able to distinguish as it aesthetically decent but with unacceptable picture quality. In Fig.~\ref{fig:morelqm}, we also show that SAQI-Local is able to distinguish \textit{white clouds (leftmost)} from \textit{over-exposed areas (rightmost)}, proving that the proposed SAQI can not only understand the goodness (meaningfulness, appealingness) of contents, but also able to detect non-typical distortions from semantic guidance, solving the challenges as mentioned in Fig.~\ref{fig:challenge}. We also upload more examples to our project website.

\subsection{Ablation Studies}
\label{sec:ablation}

In the ablation studies, we discuss the effects of different quality indices: Semantic Affinity, Spatial Naturalness and Temporal Naturalness, on either natural photo-realistic datasets (Sec.~\ref{sec:ablsi}) and YouTube-UGC (Sec.~\ref{sec:ablsytugc}). We then discuss the effects of our aggregation strategy (Sec.~\ref{sec:ablas}). We also evaluate the effects of different prompt pairs and the proposed multi-prompt aggregation (Sec.~\ref{sec:ablprompt}). %Moreover, we evaluate the effectiveness of different variants on fine-tuning (Sec.~\ref{sec:ablft}).

\begin{table*}[]
\caption{Ablation Studies (I): effects of different indices in the proposed BVQI and BVQI-Local, on three natural video datasets.}
\vspace{-6pt}
\setlength\tabcolsep{8pt}
\renewcommand\arraystretch{1.16} 
\resizebox{\textwidth}{!}{
\begin{tabular}{cc:cc|ccc|ccc|ccc}
\hline
 \multicolumn{4}{c|}{Different Quality Indices}                    & \multicolumn{3}{c|}{LIVE-VQC} & \multicolumn{3}{c|}{KoNViD-1k} & \multicolumn{3}{c}{CVD2014} \\ \hline

$\mathrm{Q}_{A, \text{Local}}$ & $\mathrm{Q}_{A}$ &$\mathrm{Q}_S$  & $\mathrm{Q}_T$ & SRCC$\uparrow$       & PLCC$\uparrow$ & KRCC$\uparrow$   & SRCC$\uparrow$       & PLCC$\uparrow$  & KRCC$\uparrow$   & SRCC$\uparrow$       & PLCC$\uparrow$    & KRCC$\uparrow$       \\
\hline
\multicolumn{13}{l}{\textit{Without Semantic Affinity Criterion:}} \\ \hdashline
&  & \cmark &   & 0.593 & 0.615 & 0.419 & 0.537 & 0.528 & 0.375 & 0.489 & 0.558 & 0.333 \\
&   &   & \cmark & 0.690 & 0.682& 0.502 & 0.577 & 0.569 & 0.404&0.482&0.498 & 0.353 \\

&   & \cmark & \cmark & 0.749 & 0.753 & 0.553 & 0.670 & 0.672 & 0.483 & 0.618 & 0.653 & 0.440 \\

\hline
\multicolumn{13}{l}{\textit{With global SAQI ($\mathrm{Q}_{A}$), highlighted row for \textbf{BVQI}:}} \\ \hdashline
& \cmark & \cmark & & 0.692 & 0.712 & 0.508 & 0.718 & 0.713 &  0.515  & 0.716  & 0.731 & 0.526 \\
& \cmark &   & \cmark & 0.767  & 0.768  & 0.568  & 0.704 & 0.699 & 0.519 & 0.708 & 0.725 & 0.502  \\
\rowcolor{lightpink} & \cmark & \cmark & \cmark & \blue{0.784} & \blue{0.794} & \blue{0.583}  & \blue{0.760} & \blue{0.760} & \blue{0.568} & \blue{0.740} & \blue{0.763} & \blue{0.542} \\ \hline
\multicolumn{13}{l}{\textit{With SAQI-Local ($\mathrm{Q}_{A,\text{Local}}$), highlighted row for \textbf{BVQI-Local}:}} \\ \hdashline

\cmark &  & \cmark & & 0.707 & 0.728 & 0.516 & 0.722 & 0.727 & 0.527 & 0.737 & 0.749 & 0.543 \\

\cmark &  &   & \cmark & 0.779 & 0.779 & 0.579 & 0.716 & 0.713 & 0.521 & 0.717 & 0.730 & 0.515\\
\rowcolor{lightpink} \cmark &  & \cmark & \cmark & \bred{0.794} & \bred{0.803} & \bred{0.594} & \bred{0.772} & \bred{0.772} & \bred{0.576} &  \bred{0.747} & \bred{0.768} & \bred{0.550} \\

\hline
\end{tabular}}
\label{tab:ablation}
\vspace{-2mm}
\end{table*}

\begin{table}[]
\caption{Ablation Studies (II): effects of different indices in the proposed BVQI and BVQI-Local on YouTube-UGC dataset.}
%\vspace{-8pt}
\centering
\renewcommand\arraystretch{1.1} 
\setlength\tabcolsep{11pt}
\resizebox{\linewidth}{!}{
\begin{tabular}{cccc|cc}
\hline
 \multicolumn{4}{c|}{Indices in BVQI/BVQI-Local}                 & \multicolumn{2}{c}{YouTube-UGC} \\ \hline
 $\mathrm{Q}_{A,\text{Local}}$ & $\mathrm{Q}_A$ &   $\mathrm{Q}_S$ & $\mathrm{Q}_T$ & SRCC$\uparrow$       & PLCC$\uparrow$ \\
\hline
& & \cmark  &   & 0.488 & 0.333 \\
& & & \cmark & 0.133 & 0.141 \\ \hdashline
& \cmark &   &   & 0.585 & {0.606} \\
& \cmark & \cmark  &   & {0.589} & 0.604 \\ & \cmark & \cmark  & \cmark  & 0.525 & 0.556 \\ \hdashline
 \cmark &  &   &   & \textbf{0.610} & \textbf{0.616} \\
\cmark & & \cmark  &   & 0.594 & 0.589 \\ \cmark & & \cmark  & \cmark  & 0.550 & 0.563 \\ 
\hline
\end{tabular}}
\vspace{-10pt}
\label{table:ugc}
\end{table}

\begin{table*}[]
\centering
\renewcommand\arraystretch{1.2} 
\setlength\tabcolsep{8pt}
\caption{Analysis on spatial downsampling during computing the SAQI, compared with variants with full-resolution frames (though strategies of \cite{clipiqa}). The variants with full-resolution frames will require much higher computation load, yet also reach much worse performance (especially on YouTube-UGC).}
%% new tabular
\resizebox{\linewidth}{!}{
\begin{tabular}{l|cc|cc|cc|cc|cc}
\hline
{Datasets}                    & \multicolumn{2}{c}{LIVE-VQC ($\leq$1080P)} & \multicolumn{2}{c}{KoNViD-1k (540P)} & \multicolumn{2}{c}{CVD2014 ($\leq$720P)} & \multicolumn{2}{c}{YouTube-UGC ($\leq$2160P)} & \multicolumn{2}{c}{LSVQ$_\text{1080P}$ (1080P)}\\ \hline
Variants & 

SRCC$\uparrow$ & PLCC$\uparrow$   & SRCC$\uparrow$& PLCC$\uparrow$  & SRCC$\uparrow$& PLCC$\uparrow$   & {~~SRCC$\uparrow$} & {~PLCC$\uparrow$}  & SRCC$\uparrow$&PLCC$\uparrow$        \\ \hline
\multicolumn{5}{l}{\textit{Part I: Variants of SAQI}:} \\ \hdashline
\textit{full-resolution} SAQI & 0.587&0.566 & 0.397&0.392 &  0.642&0.661 & {~~0.324}&{~0.307} & 0.334 & 0.308  \\
\textbf{{SAQI} (Ours)} & \blue{0.629}&\blue{0.638} & \blue{0.609}&\blue{0.602} & \blue{0.686}&\blue{0.693} & \blue{~~0.585}&\blue{~0.606}&\blue{0.527}&\blue{0.529}   \\
\textit{- improvements} & 7.2\%&12.7\% & 53.4\%&53.6\% &  6.2\%&4.4\% &{~~80.6\%}&{~97.4\%}&57.7\%&71.7\% \\ \hdashline
\multicolumn{4}{l}{\textit{Part II: Variants of SAQI-Local}:} \\ \hdashline
\textit{full-resolution} SAQI-Local & 0.629&0.607 & 0.420&0.408 & 0.543&0.575&{~~0.344}&{~0.317}&0.361&0.330 \\
\textbf{{SAQI-Local} (Ours)} & \bred{0.651}&\bred{0.663} & \bred{0.622}&\bred{0.629} & \bred{0.734}&\bred{0.731} & \bred{~~0.610}&\bred{~0.616}&\bred{0.546}&\bred{0.552}  \\
\textit{- improvements} & 3.5\%&9.2\% & 48.1\%&54.1\% &  35.2\%&27.1\% & {~~77.3\%} & {~94.3\%} & 51.2\% & 67.3\% \\
\hline

\hline
\end{tabular}}
\vspace{-2pt}
    \label{tab:downsample}
\end{table*}

\begin{table}[]
\caption{Ablation Studies (III): comparison of different alignment and aggregation strategies in the proposed BVQI quality index.}
%\vspace{-8pt}
\renewcommand\arraystretch{1.22} 
\setlength\tabcolsep{4.5pt}
\resizebox{\linewidth}{!}{
\begin{tabular}{c|c|c|c}
\hline
{\textbf{Aggregation}}                    & {LIVE-VQC} & {KoNViD-1k} & {CVD2014} \\ \hline
Metric & SRCC$\uparrow$/PLCC$\uparrow$   & SRCC$\uparrow$/PLCC$\uparrow$   & SRCC$\uparrow$/PLCC$\uparrow$         \\ \hline
\textit{Direct Addition} & 0.760/0.750 & 0.675/0.660 & 0.664/0.699 \\
\textit{Linear} + \textit{Addition} & 0.776/0.760 & 0.720/0.710  & 0.700/0.729 \\ \hdashline
 \textit{Sigmoid} + \textit{Multiplication} &  0.773/0.729 & 0.710/0.679 & 0.692/0.661 \\
\hdashline
\rowcolor{lightpink} \textit{Sigmoid} + \textit{Addition} & \bred{0.784}/\bred{0.794} & \bred{0.760}/\bred{0.760} & \bred{0.740}/\bred{0.763} \\
\hline
\end{tabular}}
\label{table:agg}
%\vspace{-10pt}
\end{table}

\subsubsection{Effects of Separate Indices on Photo-Realistic Datasets}
\label{sec:ablsi}
During evaluation on the effects of separate indices, we divide the four datasets into two parts: for the first part, we categorize the LIVE-VQC, KoNViD-1k and CVD2014 as \textbf{natural datasets}, as they do not contain computer-generated contents, or movie-like edited and stitched videos. We list the results of different settings in Tab.~\ref{tab:ablation}, where all three indices contribute notably to the final accuracy of the proposed BVQI, proving that the semantic-related quality issues, traditional spatial distortions and temporal distortions are all important to building an robust estimation on human quality perception. Specifically, in CVD2014, where videos only have authentic distortions during capturing, the Semantic Affinity ($\mathrm{Q}_A$) index shows has largest contribution; in LIVE-VQC, the dataset commonly-agreed with most temporal distortions, the Temporal Naturalness ($\mathrm{Q}_T$) index contributes most to the overall accuracy. The difference between results in diverse datasets by side validates our aforementioned claims on the separate quality concerns of the three different quality indices.

\subsubsection{Effects of Separate Indices on YouTube-UGC}
\label{sec:ablsytugc}

In YouTube-UGC, as shown in Tab.~\ref{table:ugc}, the Spatial Naturalness index cannot improve the final performance of the BVQI, where the Temporal Naturalness index even lead to 8\% performance drop. As YouTube-UGC are all long-duration (20-second) videos and almost every videos is made up of multiple scenes, we suspect this performance degradation might come from the during scene transition, where the temporal curvature is very large but do not lead to degraded quality. In our future works, we consider detecting scene transition in videos and only compute the index within the same scene. %In this stage, for fairer comparison, we still report the overall BVQI performance in the YouTube-UGC dataset, \textit{\textbf{but we recommend to use the only Semantic Affinity Quality Index $\mathrm{Q}_A$}} on professionally post-processed videos.

\subsubsection{Effects of Aggregation Strategies}
\label{sec:ablas}

We evaluate the effects of aggregation strategies in Tab.~\ref{table:agg}, by comparing with different rescaling strategies (\textit{Linear} denotes Gaussian Noramlization only, and \textit{Sigmoid} denotes Gaussian followed by Sigmoid Rescaling) and different fusion strategies (\textit{addition($+$) or multiplication($\times$)}). The results have demonstrated that the both gaussian normalization and sigmoid rescaling contributes to the final performance of aggregated index, and \textit{addition} is better than \textit{multiplication}.

\begin{table*}[]
\caption{Ablation Studies (IV): effects of different text prompts and multi-prompt aggregation in \textbf{BVQI} and \textbf{BVQI-Local}.}
\vspace{-6pt}
\renewcommand\arraystretch{1.22} 
\setlength\tabcolsep{6pt}
\resizebox{\linewidth}{!}{
\begin{tabular}{l|c|c|c|c|c|c|c}
\hline
\textit{Variants of} \textbf{BVQI} & \multicolumn{3}{c|}{Overall Performance of BVQI} & \multicolumn{4}{c}{Performance of SAQI Only} \\
\hline
{Dataset}                    & {LIVE-VQC} & {KoNViD-1k} & {CVD2014} & {LIVE-VQC} & {KoNViD-1k} & {CVD2014} & {YouTube-UGC} \\ \hline
\textbf{Prompt Pairs} & SRCC$\uparrow$/PLCC$\uparrow$   & SRCC$\uparrow$/PLCC$\uparrow$   & SRCC$\uparrow$/PLCC$\uparrow$  &SRCC$\uparrow$/PLCC$\uparrow$   & SRCC$\uparrow$/PLCC$\uparrow$   & SRCC$\uparrow$/PLCC$\uparrow$     & SRCC$\uparrow$/PLCC$\uparrow$           \\ \hline
\textbf{(a)} \textit{a [high $\leftrightarrow$low] quality photo}   & 0.768/0.775 & 0.725/0.725 & 0.738/0.757 &0.560/0.575&0.477/0.472&\bred{0.728}/\bred{0.729}&0.539/0.564\\
\textbf{(b)} \textit{a [good$\leftrightarrow$bad] photo} & 0.778/0.785 & 0.727/0.727 & 0.653/0.686 &0.608/0.581&0.586/0.551&0.507/0.512&0.473/0.458 \\ 
%\textbf{(c)}  &  0.773/0.729 & 0.710/0.679 & 0.692/0.661 \\
\hdashline
\rowcolor{lightpink} \textbf{(a)}+\textbf{(b)} \textit{Aggregated} & \bred{0.784}/\bred{0.794} & \bred{0.760}/\bred{0.760} & \bred{0.740}/\bred{0.763} &\bred{0.629}/\bred{0.638}&\bred{0.609}/\bred{0.602}&0.686/0.693&\bred{0.585}/\bred{0.606} \\
\hline
\end{tabular}}
\label{table:prompt}
%% new tabular
\resizebox{\linewidth}{!}{
\begin{tabular}{l|c|c|c|c|c|c|c}
\textit{Variants of} \textbf{BVQI-Local} & \multicolumn{3}{c|}{Overall Performance of BVQI-Local} & \multicolumn{4}{c}{Performance of Semantic Affinity Quality Localizer Only} \\
\hline
{Dataset}                    & {LIVE-VQC} & {KoNViD-1k} & {CVD2014} & {LIVE-VQC} & {KoNViD-1k} & {CVD2014} & {YouTube-UGC} \\ \hline
\textbf{Prompt Pairs} & SRCC$\uparrow$/PLCC$\uparrow$   & SRCC$\uparrow$/PLCC$\uparrow$   & SRCC$\uparrow$/PLCC$\uparrow$  &SRCC$\uparrow$/PLCC$\uparrow$   & SRCC$\uparrow$/PLCC$\uparrow$   & SRCC$\uparrow$/PLCC$\uparrow$     & SRCC$\uparrow$/PLCC$\uparrow$           \\ \hline
\textbf{(a)} \textit{a [high $\leftrightarrow$low] quality photo}   & 0.787/0.788 & 0.743/0.742 & \bred{0.768}/\bred{0.782} &0.590/0.581&0.492/0.491&0.725/0.727&0.581/0.571\\
\textbf{(b)} \textit{a [good$\leftrightarrow$bad] photo} & 0.783/0.795 & 0.746/0.749 & 0.658/0.689 &0.612/0.631&0.575/0.578&0.508/0.527&0.467/0.480 \\ 
%\textbf{(c)}  &  0.773/0.729 & 0.710/0.679 & 0.692/0.661 \\
\hdashline
\rowcolor{lightpink} \textbf{(a)}+\textbf{(b)} \textit{Aggregated} & \bred{0.794}/\bred{0.803} & \bred{0.772}/\bred{0.772} & {0.747}/{0.768} &\bred{0.651}/\bred{0.663}&\bred{0.622}/\bred{0.629}&\bred{0.734}/\bred{0.731}&\bred{0.610}/\bred{0.616} \\
\hline
\end{tabular}}
\label{table:promptlocal}
%\vspace{-14pt}
\end{table*}

\begin{table*}
\footnotesize
\caption{Ablation studies (VI): Performance of different variants for the proposed efficient fine-tuning.}\label{tab:ablft}
\setlength\tabcolsep{5pt}
\renewcommand\arraystretch{1.28}
\footnotesize
\centering
%\vspace{-8pt}
\resizebox{\textwidth}{!}{\begin{tabular}{l|c|cc|cc|cc|cc}
\hline
\textbf{Dataset} & Trainable Parameters    & \multicolumn{2}{c|}{LIVE-VQC}   & \multicolumn{2}{c|}{KoNViD-1k}        & \multicolumn{2}{c|}{YouTube-UGC}     &  \multicolumn{2}{c}{CVD2014}           \\ \hline
Variants
   &  ($\downarrow$)    & SRCC$\uparrow$   & PLCC$\uparrow$      & SRCC$\uparrow$   & PLCC$\uparrow$             & SRCC$\uparrow$   & PLCC$\uparrow$                   &SRCC$\uparrow$  & PLCC$\uparrow$                                 \\ \hline
   \gray{\textit{Zero-shot} \textbf{BVQI-Local}} & \gray{0} & \gray{0.794} & \gray{0.803} & \gray{0.772} & \gray{0.772} & \gray{0.550} & \gray{0.563} & \gray{0.747} & \gray{0.768} \\ \hdashline
   \multicolumn{10}{l}{\textit{Group 1: Variants \textbf{without} Implicit Prompt (IP)}} \\ \hdashline
   Only Optimize Final Weights ($\mathbf{W}_\mathrm{fusion}$) & 4 & 0.800 & 0.822 & 0.774 & 0.776 & 0.650 & 0.651 & 0.789 & 0.801 \\
Directly Optimize $f_t^T$ & 4,100 & 0.828 & 0.839 & 0.822 & 0.827 & 0.798 & 0.790 & 0.866 & 0.873 \\ 
%Fine-tuning the $\mathrm{E}_t$ (CLIP-Text) & 37,828,612 \\

%\textbf{\textit{Contextual Prompt}} (CP) + Fine-tuning the $\mathrm{E}_t$ & 37,844,996 \\ 

\hdashline
 \textbf{BVQI-Local} + \textbf{\textit{CP}} (Full, Ours) & 2,052 & 0.832 & 0.844 & 0.827 & 0.831 & 0.808 & {0.803} & 0.871 & {0.877} \\  \hdashline
\multicolumn{10}{l}{\textit{Group 2: Variants \textbf{with} Implicit Prompt (IP)}} \\ \hdashline
Only \textbf{\textit{Implicit Prompt (IP)}} $\mathrm{Q}_A^\text{Implicit}$ & 65,600 & 0.791 & 0.803 & 0.802 & 0.807 & 0.789 & 0.778 & 0.837 & 0.853\\
%\textbf{BVQI-Local}  + \textbf{\textit{IP}} & 65,565 & \\
\textbf{BVQI-Local} + \textbf{\textit{CP}} + Linear on Visual Features & 9,221 & 0.838 & 0.849 & 0.829 & 0.833 & 0.803 & 0.801 & 0.873 & 0.880 \\ \hdashline 
 \textbf{BVQI-Local} + \textbf{\textit{CP}} + \textbf{\textit{IP}} (Full, Ours) & 67,653 & \bred{0.840} & \bred{0.850} & \bred{0.833} & \bred{0.834} & \bred{0.816} & \bred{0.804} & \bred{0.876} & \bred{0.882} \\ \hline

\end{tabular}}
%\vspace{-13pt}
\end{table*}

\subsubsection{Effects of Downsampling}

\begin{table}[h]
\renewcommand\arraystretch{1.28} 
\setlength\tabcolsep{4.5pt}
\caption{Ablation Studies (V): Results of other prompts for SAQI-Local. All prompts are in form ``\textit{a [DESCRIPTION] photo}'' with \textit{DESCRIPTION} listed below. While \textbf{(a)/(b)} as adopted by SAQI/SAQI-Local reach better performance, the differed results on different datasets for other more concrete prompts (\textbf{(c)-(g)}) suggest the differences of datasets.}
%% new tabular
\resizebox{\linewidth}{!}{
\begin{tabular}{l|c|c|c}
\hline
 & \multicolumn{3}{c}{Performance of respective SAQI-Local} \\
\hline
{Dataset}                    & {LIVE-VQC} & {KoNViD-1k} & {CVD2014} \\ \hline
\textbf{Description Pairs} & 

SRCC$\uparrow$/PLCC$\uparrow$   & SRCC$\uparrow$/PLCC$\uparrow$   & SRCC$\uparrow$/PLCC$\uparrow$         \\ \hline
\multicolumn{4}{l}{\textbf{Perceptual-level Prompts used in SAQI:}} \\ \hdashline
\textbf{(a)} \textit{[high$\leftrightarrow$low] quality} & 0.590/0.581 & 0.492/0.491 & \bred{0.725}/\bred{0.727} \\
\textbf{(b)} \textit{[good $\leftrightarrow$bad]} & \bred{0.612}/\bred{0.631} & \bred{0.575}/\bred{0.578} & 0.508/0.527 \\
\hdashline
\multicolumn{4}{l}{\textbf{More Concrete (Unused) Prompts:}} \\ \hline
\textbf{(c)} \textit{[sharp$\leftrightarrow$fuzzy]} &  0.513/0.518 & 0.537/0.535 & 0.473/0.492  \\
\textbf{(d)} \textit{[noise-free$\leftrightarrow$noisy]} & 0.202/0.231 & 0.345/0.346 & 0.447/0.470\\ 
\textbf{(e)} \textit{[pristine$\leftrightarrow$distorted]} &0.368/0.366 & 0.373/0.380 & 0.281/0.300  \\
\textbf{(f)} \textit{[lossless$\leftrightarrow$lossy]} & 0.384/0.401 & 0.360/0.360 & 0.446/0.478 \\
\textbf{(g)} \textit{[pleasant$\leftrightarrow$annoying]} & 0.377/0.389 & 0.406/0.410 & 0.041/0.052 \\
\hline
\end{tabular}}
    \label{tab:extprompt}
    \vspace{-10pt}
\end{table}

In our proposed SAQI and SAQI-Local indices, we have implemented spatial and temporal downsampling (Sec.~\ref{sec:downsample}) to focus on semantic information of videos before feeding them to the visual backbone of CLIP. This approach stands in contrast to Wang \textit{et al.}\cite{clipiqa}, who have recently proposed an IQA method that removes the positional embedding in the Attention Pooling layer and feeds full-resolution frames as inputs. Our experiments, detailed in Tab.\ref{tab:downsample}, demonstrate that retaining the original resolution for the semantic index is suboptimal for VQA. The original-resolution variants are \textbf{\textit{neither efficient}} as it requires up around 10$\times$ running time compared to the proposed downsampled SAQI or SAQI-Local, \textbf{nor effective} as it results in notably worse performance than SAQI/SAQI-Local; moreover, the higher the original video resolutions are, the larger the performance gap between our SAQI-Local and full-resolution variants. As the downsampling technique are actually compromising low-level quality perception, the improved performance of our approach can be attributed to its enhanced ability to perceive semantic-related information.

\subsubsection{Effects of Text Prompt Pairs}
\label{sec:ablprompt}

In Tab.~\ref{table:prompt}, we discuss the effects of different text antonym pairs as $T_{+}$ and $T_{-}$ in Eq.~\ref{eq:ad}. We notice that \textit{a [high$\leftrightarrow$low] quality photo} can achieve very good performance on CVD2014 either for BVQI or the BVQI-Local, where the content diversity can be neglected and the major concern during the quality ratings is about {authentic distortions} (\textit{blurs, white balance, exposure, etc}). For LIVE-VQC and KoNViD-1k (with diverse contents), however, the \textit{a [good$\leftrightarrow$bad] photo} prompt shows higher accuracy. Specifically, in KoNViD-1k, the \textit{good/bad} pair reaches \textbf{10\%} higher correlation with human opinions than \textit{high/low quality} pair, suggesting that the subjective quality concern on this dataset might also be more prone to photo aesthetics or semantic preferences. The results suggests that different datasets have different quality concerns, while aggregating two antonym pairs can result in stable improvements for overall performance in all datasets, proving the effectiveness of the proposed multi-prompt aggregation strategy.

To further explore the quality concerns of different VQA datasets, we choose five more concrete pairs and evaluate their prompt-specific result in Tab.~\ref{tab:extprompt} (while the qualitative results for them are illustrated in Fig.~\ref{fig:lqm}). We notice that prompts related to more concrete distortion description (such as \textit{lossless$\leftrightarrow$lossy}, \textit{clean$\leftrightarrow$noisy}) are more effective on CVD2014 dataset, while these are distortions explicitly captured in this dataset. More interestingly, we find out that in LIVE-VQC and KoNViD-1k (user-generated-content datasets), the content appealingness (\textit{pleasant$\leftrightarrow$annoying}) significantly affect quality opinions, but in CVD2014 with only concern on distortions, the pair shows \textbf{almost no correlation} with human opinions. These concrete prompts further help us to investigate the mechanism behind human quality perception.

\subsection{Ablation Studies on Fine-tuning}
\label{sec:ablft}

\subsubsection{Variants of CP\&IP} We discuss the variants of contextual prompt (CP) and implicit prompt (IP). First, we evaluate the performance of only optimize the final linear weighted fusion (Sec.~\ref{sec:lwf}). The results of this variant could improve from the zero-shot version, suggesting that our aformentioned cliam that different datasets might consider different dimensions with different weights is reasonable. Still, it is much less accurate than variants with optimizable text prompts, while optimizing the Contextual Prompts show better performance than directly optimize the output text features, showing that the linguistic structural information via the fixed parts (\textit{a [DESCRIPTION] photo}) is still important. The implicit prompt with a MLP also shows better accuracy than a single linear layer, while only using the $\mathrm{Q}_A^\text{Implicit}$ instead of considering other parts in BVQI-Local shows much worse performance, suggesting that the handcraft naturalness indices are still notably useful in the fine-tuned version. In a word, the proposed fine-tuning scheme is both efficient and effective across different datasets.

\subsubsection{Length of Contextual Prompt}

\begin{figure}
    \centering
\includegraphics[width=\linewidth]{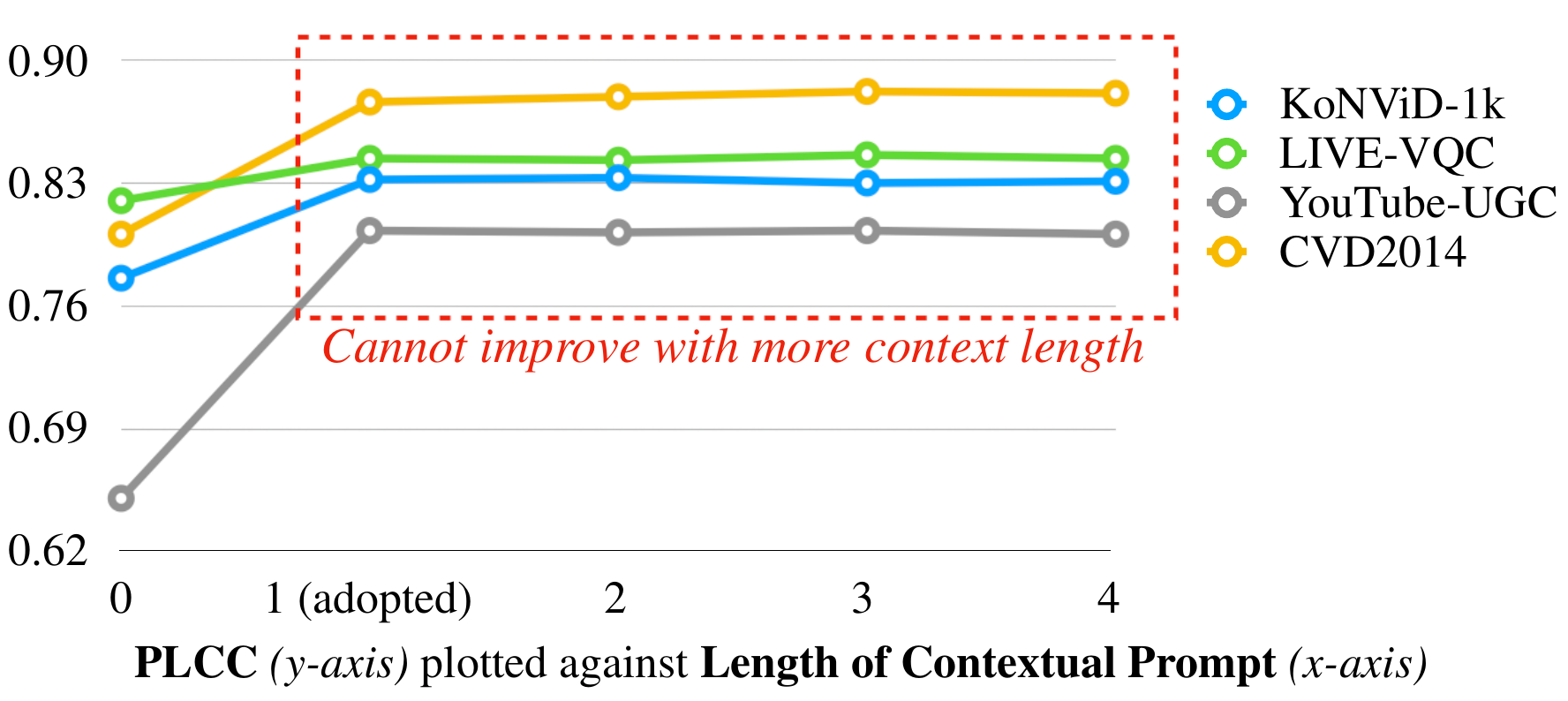}
    \vspace{-12pt}
    \caption{\textbf{PLCC} (linear correlation) result plotted against the length of contextual prompt (Sec.~\ref{sec:cp}), showing that one token (initialized as one word \textit{``X"}) as the contextual propmt is enough for fine-tuning. }
    \label{fig:ctxlen}
    \vspace{-12pt}
\end{figure}

In Fig.~\ref{fig:ctxlen}, we discuss whether the length of contextual prompt (CP) will affect the final performance. As shown in the plot, increasing the length of optimizable contexts to $>1$ will not to improve the final performance on any VQA dataset. In general, the VQA task shows much faster ``context saturation" than high-level visual tasks~\cite{coop}, which typically need 4 and more contextual prompts to reach optimal performance. This might be due to the limited data scale and relatively more simple task setting (only positive and negative classification is needed). 

\section{Conclusion and Future Works}

This paper introduces a series of zero-shot video quality indices, BVQI and BVQI-Local, which are designed to robustly assess video quality in-the-wild without training from human-labelled quality opinions. The indices combine the CLIP-based text-prompted semantic affinity quality index (SAQI) with traditional technical metrics on spatial and temporal dimensions. The proposed indices show unprecedented performance among zero-shot video quality indices. Additionally, the paper proposes a parameter-efficient fine-tuning scheme for BVQI-Local that outperforms existing training-based video quality assessment approaches, and demonstrates better robustness and competitive training speed. The fine-tuning scheme is also practical for real-world scenarios with limited quality opinions. The proposed methods can be used as reliable and effective metrics in related video research such as \textit{restoration}, \textit{generation}, and \textit{enhancement}, and potentially contribute to real-world applications such as \textit{video recommendation}.

In the future, we aim to unify the handcrafted parts of BVQI-Local, the spatial and temporal naturalness indices, into the language-vision-based SAQI. However, there are still challenges to overcome, including improving the low-level sensitivity of vision-language foundation models, modeling temporal relations (especially short-range temporal distortions) upon existing vision-language models, and improving efficiency of branches with original resolution inputs, which focus on perception of spatial technical distortions. Once these challenges are addressed, the next level of BVQI will be a stronger and integrated vision-language-based model with highly competitive robustness and efficiency.

{
\bibliographystyle{IEEEtran}
\bibliography{egbib}
}

\end{document}